%% file: number_11.tex
\documentclass[12pt]{article}

\input preamble.tex

\title{On the Effect of Bit-Level Parameter Perturbations in Machine Learning and Deep Learning Models}

\author{Akanksha Raghapur\footnotemark[1]\ \ \ 
Mark Stamp\footnotemark[1]\,\,\footnotemark[2]}

\begin{document}

\symbolfootnotetext[1]{Department of Computer Science, San Jose State University}
\symbolfootnotetext[2]{mark.stamp$@$sjsu.edu}

\maketitle

\abstract
In this chapter, we investigate how classical machine learning models 
respond to small, targeted modifications in their parameters. 
We compare and contrast these results to analogous experiments on deep learning
models. For classical learning models, we consider Hidden Markov Models (HMM) and 
Support Vector Machines (SVM), and for comparison, we conduct analogous experiments
involving Multilayer Perceptrons (MLP) and Long Short-Term Memory (LSTM) networks.
When applied to the Drebin Android malware dataset, our results show 
that classical models are brittle, in the sense
that a limited set of selected parameters can have a dramatic effect on model behavior.
In a related set of experiments, we investigate the steganographic capacity of these same learning models,
that is, the proportion of bits in model parameters that can be overwritten without
having a significant adverse affect on a model.
We find that classical models offer limited 
steganographic capacity due to their compact, parameter-efficient, and relatively 
sensitive parameter structure. In contrast, neural networks are parameter-redundant, enabling 
higher steganographic capacity, where modifications can be distributed across many parameters 
with minimal impact on performance.
These results highlight differences in how classical and neural models respond to parameter 
changes, with clear implications for both robustness and hidden information embedding. Overall, 
this work provides a framework for understanding parameter sensitivity and steganographic capacity 
across different classes of learning models.

\bigskip

\noindent \textbf{Keywords}: Parameter Perturbation $\cdot$ Bit-Level Attacks $\cdot$ Steganography $\cdot$ 
Neural Networks $\cdot$ Machine Learning $\cdot$ Deep Learning $\cdot$
Hidden Markov Model $\cdot$ Support Vector Machine $\cdot$ Multilayer Perceptron $\cdot$ Long Short-Term Memory

\section{Introduction}

Machine learning systems are increasingly deployed in security-critical environments, 
such as malware detection and network intrusion analysis. As these systems 
become integral to real-world decisions, understanding their robustness to adversarial manipulation 
is an important research concern. While adversarial perturbations 
applied to inputs have been extensively studied in the deep learning context, much less attention 
has been given to attacks that target a model’s parameters. Yet model parameters often reside 
in memory as mutable binary values, making them susceptible to low-level faults or intentional 
bit-level manipulations~\cite{Qian2023ASO}.

Recent work has shown that even highly accurate deep neural networks can be catastrophically 
degraded by modifying a small number of bits of their weights. For example, 
Rakin et al.~\cite{DBLP:journals/corr/abs-1903-12269} demonstrate that flipping as few as~13 bits 
out of~93 million parameters in a quantized ResNet-18 model is sufficient to collapse its accuracy 
on the ImageNet dataset. 
This Bit-Flip Attack (BFA), which uses a 
Progressive Bit Search (PBS) algorithm, highlights a surprisingly fragile aspect of deep learning models,
namely, the vulnerability of the parameter space itself.

This line of model robustness research has focused almost entirely on 
Deep Neural Networks (DNN). 
Classical machine learning models, such as Hidden Markov Models (HMM) and 
Support Vector Machines (SVM), are widely used in domains where interpretability, 
low computational cost, and data scarcity make DNNs impractical. Unlike deep networks, 
these models have explicit, interpretable parameters such as transition probabilities, 
emission distributions, and support vectors. Yet, despite their continued real-world 
relevance~\cite{10.1371/journal.pone.0327476}, the robustness  of classical learning models
under bit-level parameter perturbations remains largely unexplored.

On the other hand, if it is possible to substantially manipulate model parameter
without significantly altering a model’s outputs, 
this represents a potential attack surface for 
steganography or covert communication~\cite{10197462, zhang2023steganographiccapacitydeeplearning}. 
For example, an adversary who can subtly encode information by modifying the least-significant bits 
of HMM probabilities may be able to embed malware, compromise selected predictions, or evade detection, 
all without modifying the training data or observed inputs.

The research gap involving classical learning models
motivates us to consider a broader investigation into parameter-space robustness across 
different model families. In addition to classical models, modern neural architectures such as 
Multilayer Perceptrons (MLP) and Long Short-Term Memory (LSTM) networks provide an important 
point of comparison. MLPs represent feedforward neural networks with dense weight matrices, 
which are similar in structure to SVM weight vectors, while LSTMs introduce sequential modeling 
capabilities analogous to HMMs, but with significantly higher representational capacity. This enables 
a controlled comparison between structurally analogous classical and neural models, 
allowing us to analyze how architectural differences influence parameter sensitivity
and redundancy.

As part of this research, we quantize model parameters into fixed-precision integer formats.
Targeted bit flips are then applied to transition and emission matrices (for HMMs) 
and to weight vectors (for SVMs). Additional experiments with proportional probability perturbations 
are used to measure intrinsic parameter sensitivity, independent of quantization. These analyses help 
distinguish parameters that are inherently important from those that appear sensitive only due to quantization, 
providing a clearer understanding of true parameter importance.

Our results show that parameter sensitivity is highly localized in classical models, where a small subset 
of parameters significantly affect model behavior. In contrast, 
the neural models that we test generally exhibit more distributed sensitivity, 
with performance degrading more gradually under 
repeated perturbations, although certain layers remain highly sensitive.\footnote{Note that our findings regarding
MLPs and LSTMs differ from those for CNNs in Rakin et al.~\cite{DBLP:journals/corr/abs-1903-12269}, where
a small number of carefully selected weights were found to have a devastating impact on the model.}
These findings provide a direct 
comparison of parameter robustness across model types and highlight key differences in their vulnerability 
to targeted parameter corruption. Furthermore, our results show that classical models exhibit relatively 
limited steganographic capacity. In classical models, some parameters can be modified without 
affecting performance, although the number of such parameters is relatively small. On the other hand,
modifications to sensitive parameters results in rapid degradation, limiting the region of the parameter space 
available for such modification.

The remainder of this chapter is organized as follows. Section~\ref{chap:background} provides background 
on bit-level parameter perturbations, PBS, and the machine learning models considered 
in this work. Section~\ref{chap:methodology} describes our methodology, including dataset preprocessing, 
parameter representation, and perturbation techniques. Section~\ref{chap:experiments} presents the results 
of our perturbation experiments involving HMMs, SVMs, MLPs, and LSTMs. Finally, 
Section~\ref{chap:conclusion} concludes this chapter and discusses potential directions 
for future research.

\section{Background}\label{chap:background}

In this section, we first consider selected examples of relevant related work. 
We then briefly introduce the classic machine learning models
that are the focus of the research presented in this chapter. 
Finally, we emphasize the motivation for the research conducted in this chapter.

\subsection{Related Work}

Robustness in machine learning typically refers to a model’s ability to maintain its predictive 
performance under perturbations. Most prior work has examined input-space robustness~\cite{Costa_2024}, 
where adversarial examples are crafted to alter a model’s prediction while keeping the input visually 
or statistically similar.

More recently, attention has shifted toward parameter-space perturbations. 
For example, Yu, Wang, and Gao~\cite{yu2023adversarial} demonstrate that small modifications to model parameters can preserve overall accuracy on clean inputs, while significantly increasing the model's susceptibility to adversarial input examples, indicating that parameters themselves form a sensitive attack surface.
Recent work has also 
explored the steganographic capacity of neural networks by developing frameworks that embed 
data into model parameters during training, particularly for convolutional neural networks
(CNNs)~\cite{YANG2023119250}.

Importantly, these parameter-space perturbations are not purely theoretical but can be replicated 
in practical settings through hardware-level fault injection \hbox{techniques}. For example, 
Rowhammer attacks exploit vulnerabilities in DRAM to induce bit flips in memory cells, 
allowing an adversary to directly modify stored data~\cite{rowham}. 
Yao et al.~\cite{DBLP:journals/corr/abs-2003-13746} 
demonstrate that such attacks can be used to induce targeted bit flips in model parameters and 
significantly degrade deep neural network performance. More recent work further shows that these 
attacks extend to modern hardware platforms, including GPUs, where bit flips in memory can lead 
to a substantial drop in model accuracy~\cite{lin2025gpuhammer}. Since machine learning model 
parameters are typically stored as binary values in memory, such faults can corrupt weights or 
probabilities without altering the input. These results show that bit-level parameter manipulation 
is a realistic threat model.

Despite their continued 
use in real-world applications, as far as the authors are aware, classical machine learning models, 
such as HMMs and SVMs, have not been examined under parameter-space perturbations. 
In particular, it is unknown whether small, targeted changes to the stored parameters of classical models 
can meaningfully alter inference or degrade performance. This gap motivates the research in this
chapter, where we analyze how classical probabilistic models behave under bit-level manipulations 
of their quantized parameters.

Bit-flip attacks operate directly on the binary representation of a model’s stored parameters, 
flipping one or more bits in their quantized form. Because many machine learning models deploy 
quantized parameters---often stored in fixed-precision two’s complement---changing even a single 
bit can alter the underlying value of a weight or probability and potentially affect model behavior.
Perhaps the most influential example in this genre is the Bit-Flip Attack (BFA) proposed by 
Rakin et al.~\cite{DBLP:journals/corr/abs-1903-12269}, which systematically identifies sensitive 
bits whose inversion causes large changes in model loss. This previous work shows that deep 
neural networks can be extremely vulnerable to such attacks---in some cases, flipping only 
a small number of carefully chosen bits is sufficient to collapse model accuracy. This highlights 
bit-level parameter manipulation as a meaningful attack surface, independent of traditional 
input-space adversarial examples.

Rakin et al.~\cite{DBLP:journals/corr/abs-1903-12269} also show that a Progressive Bit Search (PBS) 
strategy can be used to identify high-impact bits by ranking candidates according to estimated 
influence, then iteratively flipping those most likely to increase the loss. 
This work demonstrates that some bits are far more influential than others.

The importance of individual bits has been further emphasized in follow-up work. 
Li et al.~\cite{li2025oneflip} find that even a single bit flip in a full-precision weight can embed 
a backdoor into a neural network, while Chitsaz et al.~\cite{chitsaz2023training} study how 
quantization ranges affect sensitivity to random or adversarial bit errors. Together, these 
works show that bit-level perturbations represent a meaningful axis of vulnerability in 
modern ML systems.

The broader literature also recognizes bit-flip perturbations as an important robustness 
concern in quantized models. For example, the aforementioned paper by
Chitsaz et al.~\cite{chitsaz2023training}
examines how quantization choices and parameter ranges influence a model’s sensitivity 
to bit errors. A related example is the Bit Flip Attack-guided Mixed-precision 
Neural Network Quantization technique~\cite{SUN2026114811}, 
which uses bit-flip vulnerability as a signal to guide per-layer precision choices. 
These studies show that bit-level parameter corruption is a relevant and active topic in 
machine learning research. 

Although these previous studies focus exclusively on deep neural networks, the underlying ideas extend 
naturally to other models with quantized parameters. This motivates us to investigate whether 
classical probabilistic models, such as HMMs and SVMs, 
exhibit similar sensitivities when their parameters are perturbed at the bit level.

\subsection{Classical Models and Bit-Level Perturbations}

In HMMs, transition and emission probabilities define the model behavior~\cite{Rabiner1989ATO}, 
making them natural targets for parameter-level perturbations. These probabilities can be quantized 
into fixed-precision values, allowing their bit representations to be directly perturbed. Since inference 
procedures depend explicitly on these probabilities, intuitively it seems plausible that
even small parameter changes may influence the resulting state sequences or output predictions.

In a similar vein, SVMs~\cite{cortes1995support} rely on learned weight vectors---specifically, support vector 
coefficients---that, once quantized, can also be modified at the bit level. Bit flips in these representations 
can alter margins or decision boundaries, providing another setting where the effect of parameter 
perturbations can be examined.

Together, HMMs and SVMs offer compact, structured parameter spaces that contrast sharply with 
the high-dimensional weight vectors of deep neural networks. This makes such classic models 
natural candidates for extending bit-flip analysis. In contrast to deep neural networks, 
in classic machine learning models, the roles of parameters are easier to interpret and 
perturbations can therefore potentially be studied with improved granularity.

For comparison, we also experiment with neural architectures. Specifically, we
consider MLP and LSTM networks. MLPs are 
feedforward neural networks with dense weight matrices~\cite{Rumelhart1986LearningRB}, 
structurally similar to SVM weight vectors but with greater representational capacity. 
LSTMs are a type of recurrent neural network~\cite{10.1162/neco.1997.9.8.1735} that is 
designed to model sequential data, and can therefore be viewed as a neural counterpart to 
HMMs~\cite{DBLP:journals/corr/abs-1907-04670}. These deep learning
models contain large numbers of parameters and exhibit redundancy, making them 
useful for studying how parameter distribution affects robustness and steganographic capacity.

\subsection{Research Gap and Motivation}

As discussed above, bit-flip attacks have been extensively explored in deep neural networks, 
while their impact on classical 
machine learning models appears to be unknown. Models like HMMs and SVMs use compact, 
interpretable parameters that are often quantized in practice, yet as far as the authors are aware,
no prior work examines how bit-level perturbations affect their behavior.

This creates a clear research gap, as it is not known whether classical models exhibit similar sensitivity to 
targeted bit flips as neural network models, or whether the structured parameter spaces of classic models
make them more resilient. The research in this chapter addresses this gap by applying controlled 
bit-level perturbations to HMMs and SVMs and evaluating how these changes influence 
model predictions and stability. For comparison, we perform analogous experiments with
related neural architectures, namely, MLP and LSTM.

\section{Methodology}\label{chap:methodology}

In this section, we describe the methodology used to evaluate the impact of bit-level and 
proportional parameter perturbations on the classical and neural machine learning models that we consider. 
Specifically, this section provides information on the dataset, feature representations, 
and model configurations, as well as the perturbation techniques applied---including 
least-significant-bit flipping, quantization, and Progressive Bit Search.

\subsection{Dataset and Preprocessing}

For our experiments, we use the Drebin Android malware dataset~\cite{drebin}. 
The dataset contains~15,036 Android 
applications, consisting of~6,412 malware samples from~179 families
and~8,624 benign samples. The malware samples are Android applications confirmed to exhibit malicious 
behavior such as stealing personal data, sending SMS messages without user consent, 
or secretly communicating with remote servers, while the benign samples are legitimate applications 
collected from official app stores. Each application is represented 
by~215 binary features, where each feature indicates the presence~(1) or absence~(0) of a particular static 
property extracted from the manifest or code (e.g., specific permissions, API calls, strings). 
The ground-truth label is provided in the ``transact'' column, 
with malware labeled as~1 and benign samples as~0.

For the HMM experiments, each sample is converted into a token sequence by collecting the names 
of all active features. Each unique feature is mapped to an integer ID, resulting in a vocabulary 
size of~215. The dataset is split using a 64/16/20 train/validation/test split (9,622 training samples,
2,406 validation samples, and 3,008 test samples), and the resulting sequences are concatenated
into a single observation stream with length annotations for training.

For the SVM experiments, the same 64/16/20 split is used. The raw~215-dimensional binary feature 
vectors are used directly for the linear SVM and ``true'' RBF SVM formulations. For the ``explicit'' RBF SVM, 
each input vector is mapped to feature space using RBF similarities,
providing an explicit feature representation of the data.
These RBF SVM architectures are discussed in more detail below.

For the MLP and LSTM experiments, the same~215-dimensional binary feature vectors are used as input. 
We first hold out~20\%\ of the data as the test set, then split the remaining~80\%\ into~80\%\ train 
and~20\%\ validation, which gives us an overall~64:16:20 split (9,622 samples for training, 2,406 for validation, 
and 3,008 for testing).
The features are standardized before training. 
The MLP is trained directly on these vectors, while
for the LSTM, each sample is reshaped into a sequence of length~1, treating the feature vector 
as a single time step. Aside from this reshaping, no additional preprocessing is applied.  We acknowledge that reshaping each sample into a sequence of length~1 limits the temporal modeling capabilities of the LSTM. However, since the Drebin dataset represents each application as a static binary feature vector with no inherent temporal ordering, this design choice is intentional: it allows the LSTM to be evaluated on the same feature representation as the other models, enabling a controlled comparison of parameter sensitivity across architectures.

\subsection{Parameter Representation}

In this section, we first consider least significant bit flipping, which is the most
straightforward approach for analyzing model parameter sensitivity.
Then we discuss quantization, whereby all model weights are reduced to~8-bit
representations. These quantized weights enable us to consider targeted techniques,
that are aimed at minimizing the number of bit changes required to corrupt a model.

To evaluate parameter sensitivity, least significant bit (LSB) flipping can be used to modify the binary 
representation of model parameters. For these experiments, parameters are represented in 
their default 32-bit precision formats, 
and model performance is evaluated after each level of perturbation.
Note that in the 32-bit IEEE-754 single-precision representation,~1 bit is used for the sign,~8 bits for the exponent, and the remaining~23 bits for the fraction (mantissa).

We also apply quantization to convert model parameters into fixed-precision representations.
In full-precision (e.g., 32-bit floating-point) representations, 
flipping certain bits, such as the exponent bits, will result in disproportionately large 
changes in parameter values. 
By quantizing parameters to an~8-bit 
representation, the range of values is constrained, with each bit manipulation having 
a more predictable and bounded effect on the corresponding parameter. 
This allows for systematic and meaningful evaluation of 
bit-level sensitivity, as well as consistent application of perturbation techniques. 
To analyze parameter sensitivity, a Progressive Bit Search technique is applied
to the quantized model parameters. 

Progressive Bit Search (PBS), as introduced by 
Rakin et al.~\cite{zhang2023steganographiccapacitydeeplearning}, 
is a targeted bit-level attack designed to identify the most vulnerable bits in a model’s parameter 
representation. This method iteratively selects and flips the bit that results in the largest degradation 
in model performance, allowing significant impact with a small number of modifications.

In the original formulation, PBS uses gradient information to estimate the sensitivity of each bit. 
The gradient of the loss with respect to the parameter representation is computed, and bits are 
ranked based on their influence. The most sensitive bit is then flipped, and the process is 
repeated iteratively.

PBS was developed to analyze deep learning models, and in this context it consists of two stages.
First, an in-layer search is performed, where candidate bits are evaluated within each layer.
Then a cross-layer search is performed, where the most impactful bit across all layers is selected. 
This combination enables efficient identification of highly influential bits.

In our research, the impact of bit flips is evaluated based on changes in model performance. 
For neural models (MLP and LSTM), gradients of the loss are used to guide the selection 
of sensitive bits. Specifically, the gradients are used to rank and filter candidate weights: 
we compute the gradient of the loss with respect to each weight, and restrict the bit-flip search to the
highest-gradient weights. Within that candidate set, bit flips are evaluated by actually applying each flip 
and measuring the resulting accuracy drop. Note that this differs slightly from the original PBS 
formulation of Rakin et al.~\cite{DBLP:journals/corr/abs-1903-12269}, 
which analytically scores each bit 
without requiring explicit evaluation.

For classical models, no gradients are available, and hence
bit flips are evaluated based on their impact on classification accuracy. While the original PBS 
formulation includes explicit in-layer and cross-layer search stages, these are not directly applicable 
to the classical models we consider, as neither HMMs nor SVMs have a layered structure. 
Therefore, a simplified iterative approach is used, where candidate bit flips are evaluated across 
model parameters and the most impactful bit is selected at each step. This allows a consistent 
application of PBS-style perturbations across different model types.

Note that our LSB flipping experiments enables a continuous analysis of how increasing levels of 
low-order perturbation affect model behavior.
In contrast, for the~8-bit quantized setting used in our PBS experiments, 
perturbations are restricted to a fixed-precision representation. 

\subsection{Model Configurations}

In this section, we describe the configurations of the various learning models used in our experiments.
Recall that we consider two classic ML models (HMM and SVM) and that for comparison, we consider
two deep learning models (MLP and LSTM).

\subsubsection{HMM}

For our HMMs, each sample is converted from its binary feature vector into a token sequence. 
Specifically, all of the~215 features that are active (i.e., $\mbox{value} = 1$) in a given
sample become the elements in its feature sequence. 

Two HMMs are trained, one on malware sequences and one on benign sequences. 
All HMMs are trained with~$N=6$ hidden
states, and hence the state transition matrices~$A_{\mal}$ and~$A_{\ben}$
are both~$6\times 6$, while the emission probability
matrices~$B_{\mal}$ and~$B_{\ben}$ are~$6\times 215$. This value was selected based on prior experiments applying HMMs to the Drebin dataset, where $N=6$ was found to provide a good balance between model complexity and classification performance.

Classification of a given sample is determined by the model that assigns the higher log-likelihood. 
Accuracy is then computed as the percentage of test samples whose predicted label matched the 
true (malware or benign) label. For \hbox{compactness}, all probability matrices use~32-bit floating 
point representation. For our PBS experiments, 
each parameter is further quantized to an~8-bit fixed-precision 
representation in the range of~$[0, 127]$, as discussed above.

\subsubsection{SVM}\label{sect:subSVM}

SVMs are evaluated in three configurations, namely,
a linear SVM, an ``explicit'' RBF SVM, and a ``true'' RBF SVM.
The linear SVM learns an explicit 215-dimensional weight vector defining a linear decision boundary, 
with classification performed based on the sign of the resulting score. This model is trained for
a maximum of~5,000 iterations.
For the regularization hyperparameter,
we tested each value~$C\in\{0.1, 1, 10\}$, 
and we found that~$C=1$ yielded the best validation accuracy,
and hence we use this value in all SVM experiments.

For our explicit RBF SVM, we follow the Nystr\"{o}m approximation method~\cite{NIPS2007_013a006f},
with each input vector mapped to a~300-dimensional embedding 
using RBF similarities to~300 randomly selected anchor points from the training set.
Each sample is then mapped to a~300-dimensional vector of RBF 
similarities to these anchors, defined as
$$
    \Phi(x)_k = \exp\left(-\gamma \|x - a_k\|^2\right),
$$
where~$a_k$ denotes the~$k^{\thth}$ anchor point and~$\gamma$ is the RBF kernel bandwidth parameter, set to the default value of~$1/d$ where~$d$ is the number of input features.
This gives us a model that captures nonlinear structure (through the RBF similarity embedding) 
while still having an explicit, quantizable weight vector, making it directly comparable to the 
standard linear SVM under PBS.

Our true RBF SVM uses the kernel trick, parameterized by the dual coefficients~$\alpha_i$
associated with each support vector. As with all models in our PBS experiments, 
these coefficients are extracted
and quantized to~8-bit signed integers by scaling to the range of~$[-127, 127]$, enabling 
bit-level analysis at the parameter level.

\subsubsection{MLP}

Our MLP is trained directly on the Drebin dataset feature vectors. 
The model includes two hidden layers 
(\texttt{fc1}: $215 \rightarrow 128$, \texttt{fc2}: $128 \rightarrow 64$)
and an output layer~$(64 \rightarrow 1)$. 
ReLU activation is used in the hidden layers and sigmoid is used in the output layer. 
Classification is performed based on a threshold of~0.5 for the output probability.

\subsubsection{LSTM}

An LSTM network is trained on the Drebin feature vectors.  
The model consists of an LSTM layer with~64 hidden units followed by an output 
layer~$(64 \rightarrow 1)$.  The LSTM processes each input sequence, 
and the final hidden state is passed to a fully connected layer. A sigmoid activation 
is used in the output layer, and classification is performed by thresholding the output probability.

\subsection{Evaluation Metrics}

Model performance is evaluated primarily using classification accuracy, 
defined as the percentage of correctly classified samples. Accuracy is used across 
all models to measure the impact of parameter perturbations.

For all models, a 64/16/20 train/validation/test split is used. The validation set is used to guide PBS bit selection, while the test set is reserved for final evaluation. By employing validation data, 
we ensure that perturbation effects are measured without affecting the test set.

\section{Experiment and Results}\label{chap:experiments}

In this section, we present the results of our parameter perturbation experiments for classical 
and neural machine learning models. The impact of bit-level and proportional 
modifications is evaluated in terms of model performance degradation and parameter 
sensitivity across HMM, SVM, MLP, and LSTM models.

\subsection{HMM Results}

Our HMM experiments serve to evaluate the impact of both bit-level and proportional 
perturbations on model performance and parameter sensitivity. Note that the baseline classification 
accuracy of our HMM model is~0.9137. 

Across all experiments, performance degradation is observed under both LSB-based perturbations 
and targeted bit-flip attacks, with the most significant impact arising from a small subset of 
highly sensitive parameters. The following subsections describe these results in detail.

\subsubsection{HMM Bit-Flip Procedure}

\begin{figure}[!htb]
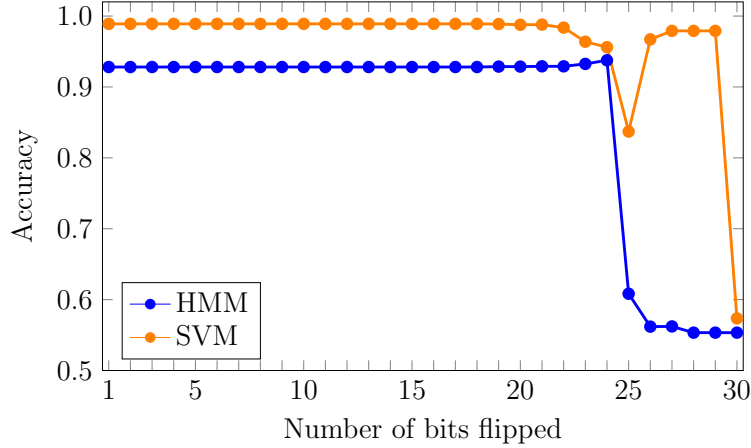

    \centering
    \input figures/lsb_hmm_svm.tex
    \caption{Effect of LSB flips on HMM and linear SVM accuracy}
    \label{fig:LSB on HMM and SVM}
\end{figure}

The least significant bits (LSB) of the emission probabilities are flipped.
For each experiment that we conduct, 
the~$k \in \{1,2,\ldots,30\}$ LSBs are flipped, the affected rows are renormalized, 
and the model is re-evaluated on the test set. This produces the accuracy curve shown 
in Figure~\ref{fig:LSB on HMM and SVM}.
The figure shows that both models maintain high accuracy when only a small number 
of LSBs are flipped, but degrade sharply beyond approximately~23-25 bits. The HMM drops 
to near random chance and remains there while the SVM shows more erratic behavior before
collapsing at~30 bits. The temporary increase in SVM accuracy observed after~26 bits is likely due to the non-monotonic nature of bit-flip perturbations, where flipping certain bits can partially counteract the effect of prior flips, resulting in transient accuracy recovery before further degradation.

\subsubsection{Progressive Bit-Flip Search on Quantized HMM}

To extend the LSB-flipping experiments, a targeted bit-flip procedure inspired by the 
PBS method of Rakin et al.~\cite{DBLP:journals/corr/abs-1903-12269} 
is applied to the~8-bit quantized HMM. After training in full precision, both the transition and 
emission matrices are quantized, and PBS is used to iteratively identify the bit whose inversion 
causes the largest decrease in model performance. 

Unlike the original PBS formulation for neural networks, which relies on analytical gradients, 
direct gradient computation is not readily available for HMMs. 
Instead, at each iteration, every candidate bit flip across the transition and emission matrices is evaluated 
by temporarily applying the flip, recomputing classification accuracy on the validation set, and recording the 
accuracy drop. The flip that causes the largest decrease in accuracy is then permanently applied, and 
the process repeats. This brute-force search effectively identifies the most impactful parameters by directly 
measuring their influence on model behavior. Average log-likelihood is also recorded after each flip as 
a diagnostic metric, but is not used to guide the search.

\begin{figure}[!htb]
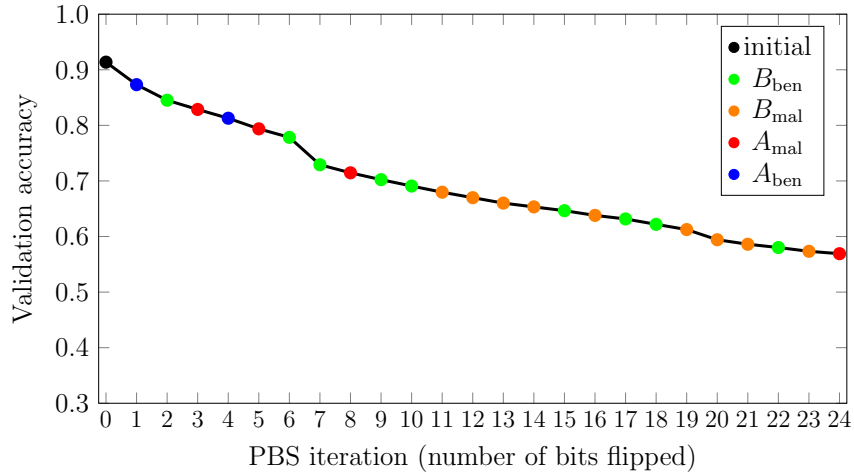

    \centering
    \input figures/pbs_hmm.tex
    \caption{Effect of PBS on HMM accuracy} 
    \label{fig:PBS on HMM}
\end{figure}

Figure~\ref{fig:PBS on HMM} shows the resulting degradation in classification accuracy as 
the number of bit flips increases.The model starts at~0.9137 accuracy and steadily declines, dropping to~0.7937 after the 
first five iterations. The degradation continues more gradually, stabilizing near~0.57 
after~24 iterations.

Beyond iteration~20, the degradation slows and the accuracy stabilizes near~57\%, 
close to random-chance performance. This suggests that once the most sensitive 
parameters are disrupted, additional bit flips have limited impact. The graph
in Figure~\ref{fig:PBS on HMM}
also indicates that perturbations are distributed across both emission and transition 
matrices throughout the attack, with both $B_{\ben}$ and $B_{\mal}$ contributing 
from early iterations.

\subsubsection{Transition Matrix Results}

Our transition matrix experiments begin with a baseline classification accuracy of~0.9137,
corresponding to the quantized model. 
After excluding the sign bit from the search space, the PBS procedure consistently selects 
mid-level bits, primarily bits~0, 2, and~5, as shown in Table~\ref{tab:pbstm}.
Recall that all HMMs have~$N=6$ hidden states and a vocabulary of size~$M=215$
is used, and hence the~$A$ matrices are~$6\times 6$, while the~$B$ matrices 
are~$6\times 215$. Thus, each~$A$ matrix has~36 parameters and each~$B$ matrix
has~1,290 parameters. With quantized~7-bit representations, 
there are~252 bits per$A$ matrix and~9,030 bits per~$B$ matrix, 
for a total of~9,282 bits per model.

\begin{table}[!htb]
\caption{PBS-selected bit flips for the quantized HMM transition matrix}\label{tab:pbstm}
\centering
\begin{adjustbox}{scale=0.85}
\begin{tabular}{c|ccccc}\toprule
\multirow{2}{*}{\textbf{Iteration}} & \multirow{2}{*}{\textbf{Matrix}} & \multirow{2}{*}{\textbf{Bit}} & \textbf{State} 
	& \multirow{2}{*}{\textbf{Accuracy}} & \multirow{2}{*}{\textbf{Drop}} \\ 
 &  &  & \textbf{transition} 
	&  &  \\ \midrule
\zz1 & $A_{\ben}$ & 5 & $4 \rightarrow 3$ & 0.8732 & 0.0370 \\
\zz3 & $A_{\mal}$ & 2 & $3 \rightarrow 4$ & 0.8286 & 0.0166 \\
\zz4 & $A_{\ben}$ & 2 & $4 \rightarrow 5$ & 0.8128 & 0.0157 \\
\zz5 & $A_{\mal}$ & 2 & $0 \rightarrow 1$ & 0.7937 & 0.0191 \\
\zz8 & $A_{\mal}$ & 5 & $2 \rightarrow 4$ & 0.7146 & 0.0145 \\
24 & $A_{\mal}$ & 0 & $0 \rightarrow 4$ & 0.5691 & 0.0043 \\ \bottomrule
\end{tabular}
\end{adjustbox}
\end{table}

The PBS procedure selects bit flips across both $A_{\mal}$ and $A_{\ben}$, 
targeting different state transitions rather than concentrating on a single entry. 
The most significant early impact comes from $A_{\ben}$ at iteration~1, with a drop of~0.0370, 
while $A_{\mal}$ hits contribute steadily through iteration~24. Overall, transition matrix 
perturbations contribute throughout the attack rather than being confined to later iterations, 
with accuracy decreasing from~0.8732 to~0.5691 over the iterations listed in 
Table~\ref{tab:pbstm}.

\subsubsection{Emission Matrix Results}

The same trend observed in Figure~\ref{fig:PBS on HMM} is reflected in the emission matrix results. 
Starting from the same baseline accuracy of~0.9137, the selected bit flips lead to a steady 
decline in performance, with the most significant drops occurring in the early iterations. 
As shown in Table~\ref{tab:pem}, perturbations occur across both the malware and benign emission matrices, with  $B_{\ben}$ 
dominating early iterations.

\begin{table}[!htb]
\caption{PBS-selected bit flips for the quantized HMM emission matrix}\label{tab:pem}
\centering
\begin{adjustbox}{scale=0.75}
\advance\tabcolsep by -2.5pt
\begin{tabular}{c|cccccc}\toprule
\textbf{Iter.} & \textbf{Matrix} & \textbf{Bit} & \textbf{Location} & \textbf{Feature} 
	& \textbf{Accuracy} & \textbf{Drop} \\ \midrule
\zz2  & $B_{\ben}$ & 6 & (0, \zz36) & \texttt{Ljavax.crypto.Cipher}                              & 0.8452 & 0.0281 \\
\zz6  & $B_{\ben}$ & 4 & (3, \zz\zz9)  & \texttt{Ljava.net.URLDecoder}                           & 0.7784 & 0.0153 \\
\zz7  & $B_{\ben}$ & 0 & (0, \zz\zz9)  & \texttt{Ljava.net.URLDecoder}                           & 0.7291 & 0.0493 \\
\zz9  & $B_{\ben}$ & 6 & (5, 113) & \texttt{GLOBAL\_SEARCH}                                      & 0.7023 & 0.0123 \\
10 & $B_{\ben}$ & 1 & (5, \zz86) & \texttt{BLUETOOTH}                                            & 0.6908 & 0.0115 \\
11 & $B_{\mal}$ & 2 & (4, 103) & \texttt{TelephonyManager.isNetworkRoaming}                     & 0.6797 & 0.0111 \\
12 & $B_{\mal}$ & 0 & (2, \zz95) & \texttt{remount}                                              & 0.6699 & 0.0098 \\
13 & $B_{\mal}$ & 1 & (0, 132) & \texttt{ACCESS\_COARSE\_LOCATION}                               & 0.6601 & 0.0098 \\
14 & $B_{\mal}$ & 0 & (2, \zz\zz1)  & \texttt{bindService}                                      & 0.6533 & 0.0068 \\
15 & $B_{\ben}$ & 6 & (3, \zz21) & \texttt{READ\_SMS}                                            & 0.6465 & 0.0068 \\
16 & $B_{\mal}$ & 2 & (4, \zz95) & \texttt{remount}                                              & 0.6380 & 0.0085 \\
17 & $B_{\ben}$ & 3 & (0, \zz41) & \texttt{WRITE\_HISTORY\_BOOKMARKS}                            & 0.6316 & 0.0064 \\
18 & $B_{\ben}$ & 2 & (0, \zz72) & \texttt{android.intent.action.PACKAGE\_REMOVED}              & 0.6219 & 0.0098 \\
19 & $B_{\mal}$ & 4 & (2, \zz52) & \texttt{android.intent.action.PACKAGE\_REPLACED}             & 0.6125 & 0.0094 \\
20 & $B_{\mal}$ & 0 & (4, \zz15) & \texttt{Landroid.content.Context.registerReceiver}           & 0.5942 & 0.0183 \\
21 & $B_{\mal}$ & 3 & (2, \zz11) & \texttt{android.telephony.SmsManager}                        & 0.5861 & 0.0081 \\
22 & $B_{\ben}$ & 1 & (0, \zz96) & \texttt{android.intent.action.ACTION\_SHUTDOWN}              & 0.5802 & 0.0060 \\
23 & $B_{\mal}$ & 0 & (2, \zz30) & \texttt{DexClassLoader}                                      & 0.5734 & 0.0068 \\ \bottomrule
\end{tabular}
\end{adjustbox}
\end{table}

The model accuracy decreases from~0.8732 to~0.7937 within the first five iterations. The 
degradation then continues more gradually, reaching approximately~0.5734 after~23 iterations, 
at which point the model has approached random-chance performance. This behavior indicates that 
a relatively small number of targeted perturbations is sufficient to significantly disrupt the model.

The selected bit flips target a mix of benign and malware model parameters. Early iterations 
concentrate on $B_{\ben}$, targeting features such as
\begin{itemize}
\item \texttt{Ljavax.crypto.Cipher}
\item \texttt{Ljava.net.URLDecoder}
\item \texttt{BLUETOOTH}
\item \texttt{GLOBAL\_SEARCH}
\end{itemize} 
Later iterations shift toward~$B_{\mal}$, targeting
\begin{itemize}
\item \texttt{remount}, 
\item \texttt{ACCESS\_COARSE\_LOCATION}
\item \texttt{DexClassLoader} 
\item \texttt{android.telephony.SmsManager} 
\end{itemize}
These features produce 
relatively large accuracy drops in the early iterations, while later iterations result in smaller 
incremental changes, indicating diminishing impact, once the most sensitive parameters 
have been disrupted.

\begin{figure}[!htb]
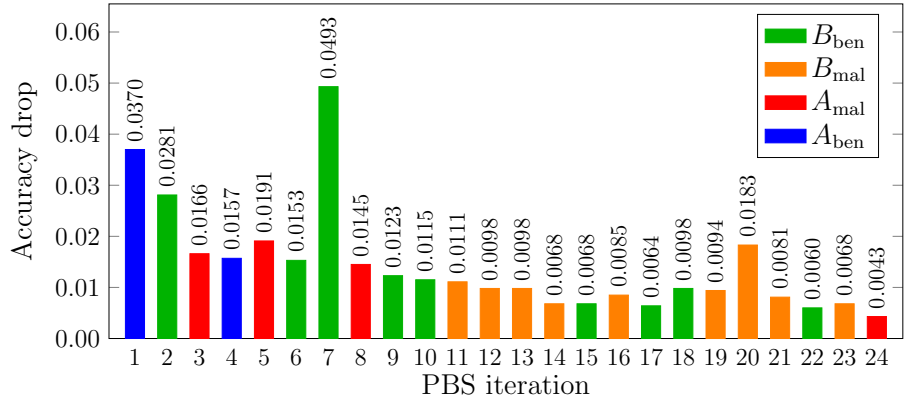

    \centering
    \input figures/bar_all.tex
    \caption{PBS-selected matrix and bit-flip impact across iterations}
    \label{fig:HMM}
\end{figure}

The feature selection timeline in Figure~\ref{fig:HMM} provides a visual representation of the results 
summarized in Tables~\ref{tab:pem} and~\ref{tab:pbstm}. Each bar corresponds to a feature selected at a 
given PBS iteration, with the color indicating the matrix, and the bar height giving the resulting accuracy drop.

Although PBS does not repeatedly select the same feature across iterations, only
a small number of distinct features are selected. Specifically, the features selected 
by PBS are the following.
\begin{itemize}
\item \texttt{Ljavax.crypto.Cipher} --- API used for cryptographic operations, commonly associated with encrypted command-and-control communication.
\item \texttt{Ljava.net.URLDecoder} --- URL decoding API, often used in obfuscated network calls; appears twice across different hidden states.
\item \texttt{GLOBAL\_SEARCH} --- System-level search permission.
\item \texttt{BLUETOOTH} --- Wireless communication permission.
\item \texttt{TelephonyManager.isNetworkRoaming} --- Checks roaming status, used in targeted attacks that behave differently based on network context.
\item \texttt{remount} --- Filesystem remount operation, a strong indicator of privilege escalation; appears twice across different hidden states.
\item \texttt{ACCESS\_COARSE\_LOCATION} --- Location access permission.
\item \texttt{bindService} --- Service binding, common in both benign and malware apps, making it influential in both HMMs.
\item \texttt{READ\_SMS} --- SMS read permission, commonly associated with malware.
\item \texttt{WRITE\_HISTORY\_BOOKMARKS} --- Browser history access permission.
\item \texttt{android.intent.action.PACKAGE\_REMOVED} and \texttt{PACKAGE\_REPLACED} --- Package lifecycle monitoring, potentially used to detect security tool installations.
\item \texttt{Landroid.content.Context.registerReceiver} --- Broadcast receiver registration, common in both malware and benign apps.
\item \texttt{android.telephony.SmsManager} --- SMS sending API, strongly associated with malware.
\item \texttt{android.intent.action.ACTION\_SHUTDOWN} --- Shutdown broadcast receiver.
\item \texttt{DexClassLoader} --- Dynamic code loading, a strong indicator of obfuscated malware behavior.
\end{itemize}
These features---which are the most sensitive to bit-flipping with respect to the Drebin dataset---fall into the 
following three categories.
\begin{enumerate}
\item Cryptographic and network obfuscation, including APIs used for encrypted or obfuscated 
communication (\texttt{Ljavax.crypto.Cipher}, \texttt{Ljava.net.URLDecoder}).
\item System-level and privilege escalation operations, indicating access to protected resources 
(\texttt{remount}, \texttt{DexClassLoader}, \texttt{TelephonyManager.isNetworkRoaming}).
\item Permissions and APIs common to both malware and benign apps, whose presence in both HMMs 
makes them disproportionately influential 
(\texttt{bindService}, \texttt{registerReceiver}, \texttt{READ\_SMS}).
\end{enumerate}
The first two categories are intuitive indicators of malware, while the third reflects parameters shared across both model types, and this intuition is supported by the research in~\cite{drebin}.

Compared to the emission matrix, the impact of transition perturbations (as discussed in the previous
section) is less dominant, suggesting that while transitions contribute to model behavior, 
their influence is secondary to emission probabilities. This behavior suggests that
while emission probabilities dominate HMM classification, a small number of transition parameters 
also play an important role. Disrupting these transitions alters the flow of probability mass across 
hidden states, causing cascading effects in sequence likelihood estimation. 
Sensitivity is distributed across both $A_{\mal}$ and $A_{\ben}$, indicating that transition 
parameters in both models contribute to classification behavior.

\subsubsection{Probability Sensitivity Analysis on HMMs}

To measure parameter importance independently of quantization, the following proportional sensitivity 
analysis is performed. For each transition and emission probability~$P_{i,j}$, the value is scaled 
by a range of percentage factors (both increases and decreases), the corresponding row is renormalized, 
and the resulting change in average log-likelihood is recorded. This analysis is applied across all 
four matrices ($A_{\mal}$, $B_{\mal}$, $A_{\ben}$, and~$B_{\ben}$). Note that 
while emission matrix entries can be 
directly mapped to feature names, transition matrix entries correspond only to hidden state 
transitions, making them less amenable to interpretation at the feature level.

Across all perturbation levels tested, only a small subset of emission probabilities produce noticeable 
changes in average log-likelihood, while the vast majority show near-zero sensitivity. 
Furthermore, the sensitive 
entries are localized to specific state-feature combinations, rather than distributed uniformly across 
the matrix. Even at larger perturbation magnitudes, the observed changes remain relatively small, 
indicating that the HMM relies on a limited set of influential parameters, with most others contributing 
minimally to the overall likelihood.

Although most emission probabilities show negligible sensitivity, a small set of features consistently 
appear among the top-ranked sensitive parameters across all perturbation levels. These represent 
the emission entries most strongly tied to the model's likelihood scoring behavior. 
Tables~\ref{tab:sensitive_emission_features} and~\ref{tab:sensitive_emission_features_benign} 
list the emission features that were consistently sensitive under~5\%, 10\%, 20\%, 50\%, 
and~75\%\ perturbations for the malware and benign HMMs, respectively.

\begin{table}[!htb]
\caption{Malware HMM emission features most sensitive across perturbations}
\centering
\begin{adjustbox}{scale=0.825}
\begin{tabular}{c | l l}
\toprule
\textbf{(row, col)} & \ \ \textbf{Feature name} & \textbf{Description} \\
\midrule
(3, \zz1)  & \ \ \texttt{INTERNET} & Network access permission \\
(5, 11) & \ \ \texttt{READ\un PHONE\un STATE} & Device ID/telephony info \\
(0, \zz9)  & \ \ \texttt{WRITE\un EXTERNAL\un STORAGE} & File-system write capability \\
(4, 10) & \ \ \texttt{HttpUriRequest} & HTTP request \\
(5, 26) & \ \ \texttt{android.intent.action.BOOT\un COMPLETED} & Persistence on device boot \\
(2, 60) & \ \ \texttt{android.telephony.SmsManager} & SMS sending/processing \\
(0, 36) & \ \ \texttt{attachInterface} & IPC system service interaction \\
(0, \zz2)  & \ \ \texttt{Binder} & Inter-process communication \\
(2, \zz8)  & \ \ \texttt{android.content.pm.PackageInfo} & Package metadata access \\
(1, 27) & \ \ \texttt{onServiceConnected} & Service binding callback \\
\bottomrule
\end{tabular}
\end{adjustbox}
\label{tab:sensitive_emission_features}
\end{table}

\begin{table}[!htb]
\caption{Benign HMM emission features most sensitive across perturbations}
\centering
\begin{adjustbox}{scale=0.725}
\begin{tabular}{c | c l}
\toprule
\textbf{(row, col)} & \textbf{Feature name} & \textbf{Description} \\
\midrule
(5, 19) & \texttt{Landroid.content.Context.registerReceiver} & Broadcast receiver registration \\
(4, 35) & \texttt{Landroid.content.Context.unregisterReceiver} & Broadcast receiver deregistration \\
(3, 28) & \texttt{ServiceConnection} & Service binding interface \\
(2, 27) & \texttt{onServiceConnected} & Service binding callback \\
(2, 21) & \texttt{android.os.Binder} & Inter-process communication \\
(0, 30) & \texttt{bindService} & Background service binding \\
(4, 10) & \texttt{HttpUriRequest} & HTTP request \\
(5, 15) & \texttt{HttpPost.init} & HTTP POST request initialization \\
(0, 36) & \texttt{attachInterface} & IPC system service interaction \\
(3, \zz9)  & \texttt{WRITE\un EXTERNAL\un STORAGE} & File-system write capability \\
\bottomrule
\end{tabular}
\end{adjustbox}
\label{tab:sensitive_emission_features_benign}
\end{table}

The malware HMM's most sensitive emission features are dominated by permissions 
and~API calls commonly associated with malicious behavior, including 
\texttt{INTERNET}, \texttt{READ\un PHONE\un STATE}, and \texttt{WRITE\un EXTERNAL\un STORAGE}, 
along with reflection APIs such as \texttt{HttpUriRequest} and IPC-related calls such 
as \texttt{attachInterface} and \texttt{Binder}. In contrast, the benign HMM's sensitivity
is concentrated in service binding and IPC mechanisms such 
as \texttt{registerReceiver}, \texttt{unregisterReceiver}, \texttt{ServiceConnection}, 
and \texttt{bindService}, reflecting the legitimate inter-process communication patterns 
that characterize benign applications.

\begin{figure}[!htb]
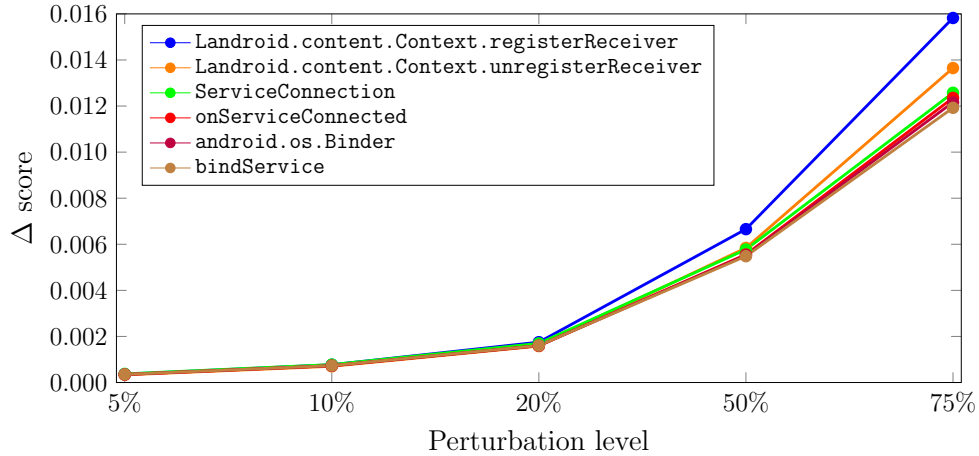

    \centering
    \input figures/sense_ben_hmm.tex
    \caption{Sensitivity of emission features for benign HMM}
    \label{fig:graph of sensitive em. features Benign}
\end{figure}

\begin{figure}[!htb]
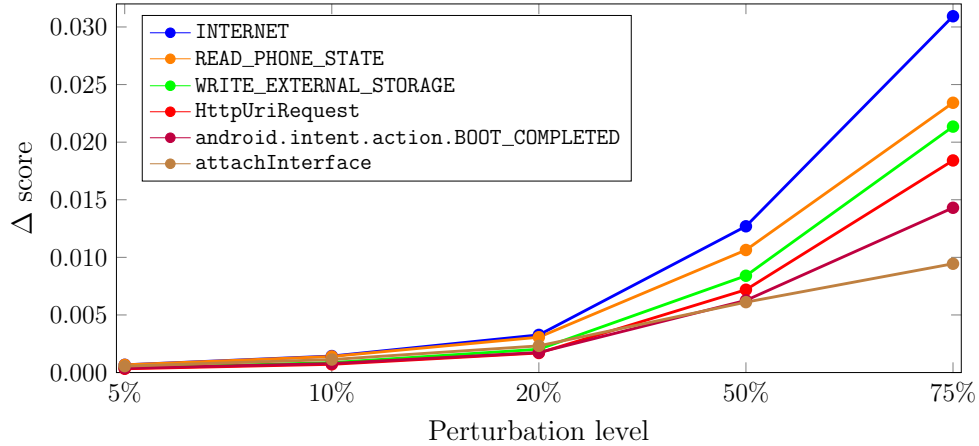

    \centering
    \input figures/sense_mal_hmm.tex
    \caption{Sensitivity of emission features for malware HMM}
    \label{fig:graph of sensitive em. features malware}
\end{figure}

Figures~\ref{fig:graph of sensitive em. features Benign} 
and~\ref{fig:graph of sensitive em. features malware} 
plot the~$\Delta$ score (i.e., the change in average log-likelihood) 
for the top six sensitive features from the benign and malware HMMs, respectively, 
across all perturbation levels. In both models, sensitivity scales nonlinearly with perturbation magnitude, 
with \texttt{INTERNET} showing the sharpest growth in the malware HMM, and 
$\mbox{\texttt{Landroid.content.Context.registerReceiver}}$ 
being the dominant feature for the benign HMM.
Features cluster 
closely at lower perturbation levels (5\%\ to~20\%) but diverge significantly at~50\%\ and~75\%, 
indicating that the most sensitive parameters become disproportionately influential under 
larger perturbations, as expected.

A comparison between the emission features identified by this proportional sensitivity analysis 
and those targeted by the PBS bit-flip attack reveals one feature that appears in both experiments 
for the malware HMM: \texttt{android.telephony.SmsManager}. 
For the benign HMM, \texttt{bindService} appears in both analyses. These overlapping features 
represent parameters that are simultaneously important to the statistical behavior of the HMM 
and fragile under quantized bit-level perturbations. Their presence in both analyses indicates that 
they should be avoided for steganographic embedding, as even minor modifications to these entries are 
likely to produce detectable changes in the model output. The remaining sensitive features identified by 
each method do not overlap, showing that PBS and proportional sensitivity analysis capture complementary 
aspects of parameter vulnerability.

\subsection{SVM Results}

Our SVM experiments evaluate the effect of bit-level perturbations on both linear and nonlinear models. 
The baseline classification accuracy is~0.9871 for the linear SVM,~0.9867 for the explicit RBF SVM, 
and~0.9907 for the true RBF SVM.

Across the perturbation experiments, the SVM models exhibit high robustness to small bit-level changes, 
with accuracy remaining stable under low-magnitude perturbations and degrading only after more 
significant modifications. The results also highlight that only a subset of parameters contribute
meaningfully to model performance, while many others have minimal impact. The following subsections 
describe the results of our bit-flip and targeted PBS experiments in detail.

\subsubsection{SVM Bit-Flip Procedure}

For the linear SVM, the LSBs~1 through~30 of each weight in the learned weight vector are flipped and 
accuracy is measured after each perturbation. The results of these experiments are summarized in 
Figure~\ref{fig:LSB on HMM and SVM} where, for comparison, 
we have also included the analogous results for our HMM experiments.

For both models, flipping the LSBs shows that accuracy remains stable through the initial range 
of perturbations and only begins to decline after more than~20 bits have been flipped. 
However, the SVM shows considerably more robustness---as compared to the 
HMM---for larger numbers of bit-flips,
This experiment clearly indicates that the SVM model can be stored in an~8-bit quantized 
format, rather than full~32-bit precision.

\subsubsection{Progressive Bit-Flip Search on Quantized SVMs}

Next, we apply the PBS procedure to our SVM model. 
Similar to the HMM setup, the model parameters are first quantized to an~8-bit representation. 
For the linear SVM, this corresponds to the weight vector, while for nonlinear SVMs, the attack 
targets either the explicit feature weights or the dual coefficients (depending on the specific SVM model
under consideration) associated with support vectors. 
At each iteration, PBS identifies the single most sensitive bit whose inversion results in the largest 
degradation in model performance. The selected bit is then permanently flipped, and the process 
is repeated over multiple iterations to progressively reduce accuracy.

For our first SVM experiment, the PBS procedure is applied to the weight vector of a linear SVM.
This experiment begins with a baseline accuracy of approximately~0.9867.

\begin{figure}[!htb]
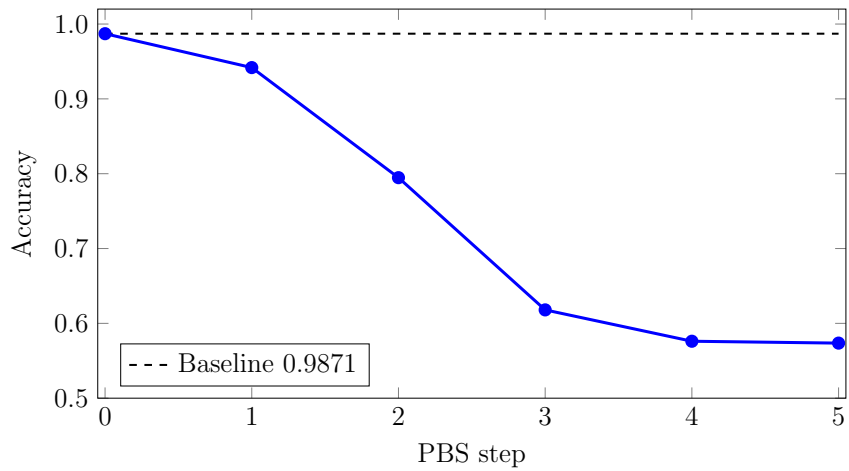

    \centering
    \input figures/pbs_linear_svm.tex
    \caption{Effect on accuracy in a linear SVM using PBS}
    \label{fig:PBS on linear SVM}
\end{figure}

After the first bit flip, 
accuracy decreases only slightly to~0.9418, followed by a sharp drop to~0.7947 after the second iteration. 
Subsequent perturbations further reduced accuracy to~0.6180, 0.5761, and~0.5736 over the first five iterations.
These results, as summarized in Figure~\ref{fig:PBS on linear SVM}, indicate that the linear SVM is highly 
sensitive to targeted perturbations in a small subset of features. In particular, modifying only 
a few critical weights is sufficient to degrade model performance to near random, demonstrating that the 
decision boundary is strongly influenced by a limited number of parameters.

To study a version of an SVM that is both nonlinear 
and has explicit, attackable weight parameters, we use an explicit RBF feature map inspired 
by the Nystr\"{o}m approximation method~\cite{NIPS2007_013a006f},
as discussed in Section~\ref{sect:subSVM}, above. 
Recall that we refer to this as an explicit RBF SVM

For our true RBF SVM, we train the model using the standard kernel formulation, 
where predictions depend on a weighted combination of support vectors rather than 
an explicit feature weight vector. The learned parameters consist of dual 
coefficients 
associated with each support vector, and a bias term.

To enable bit-level analysis on our true RBF SVM, 
the dual coefficients are extracted and quantized into~8-bit signed integers. 
This is accomplished by scaling the coefficients to the range~$[-127,127]$, followed by rounding and casting 
\hbox{to~8-bit} \hbox{integers}. During evaluation, these quantized values are de-quantized back to floating-point 
and used to compute predictions via the RBF kernel.
The PBS procedure is applied directly to the quantized dual 
coefficients. At each iteration, individual bits (excluding the sign bit) of each coefficient are flipped, 
and the resulting change in classification accuracy is evaluated. The bit flip that produces the
largest decrease in accuracy is selected and permanently applied, and this 
process is repeated iteratively.

Unlike the linear and explicit RBF SVM formulations, where parameters correspond directly to 
feature weights, the dual coefficients in the true RBF SVM are tied to specific support vectors. 
Thus, each bit flip effectively alters the influence of a particular training sample on the 
decision boundary.

\begin{figure}[!htb]
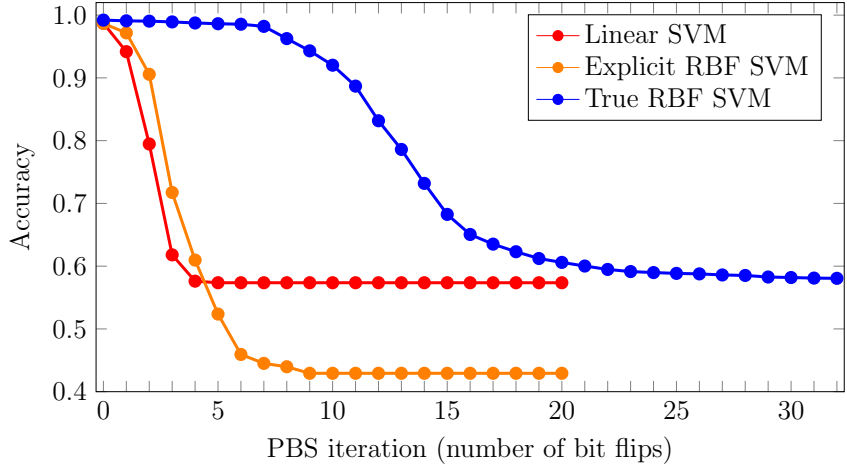

    \centering
    \input figures/pbs_svm_all.tex
    \caption{Accuracy degradation across SVM formulations}
    \label{fig:accuracy svm all}
\end{figure}

Figure~\ref{fig:accuracy svm all} gives our PBS results for all three SVM models.
The observed robustness of the true RBF SVM suggests a higher steganographic capacity compared 
to the linear and explicit RBF formulations. Since the model maintains near-baseline accuracy across 
many bit-flip iterations, a larger number of parameter bits can be modified without significantly 
affecting performance. This indicates that the dual coefficients contain substantial redundancy, 
allowing information to be embedded in their lower-order bits with minimal impact on classification 
accuracy. In contrast, the rapid degradation observed in the linear SVM implies low capacity, 
as even a small number of bit modifications disrupts the decision boundary. The gradual decline 
in the true RBF SVM highlights that its decision function is distributed across many support vectors, 
enabling more flexible and less detectable parameter modifications. Overall, the results in 
Figure~\ref{fig:accuracy svm all} suggest that kernel-based SVMs provide a more favorable 
structure for steganographic embedding than models with highly concentrated parameter sensitivity.

However, it is important to note that the true RBF SVM may not 
be directly comparable to the other two SVM formulations, 
as its parameters are dual coefficients tied to support vectors, rather than explicit feature weights. 
The comparison here is therefore in terms of observed robustness to bit-level perturbations, 
rather than a structural equivalence of the parameter spaces.

\begin{figure}[!htb]
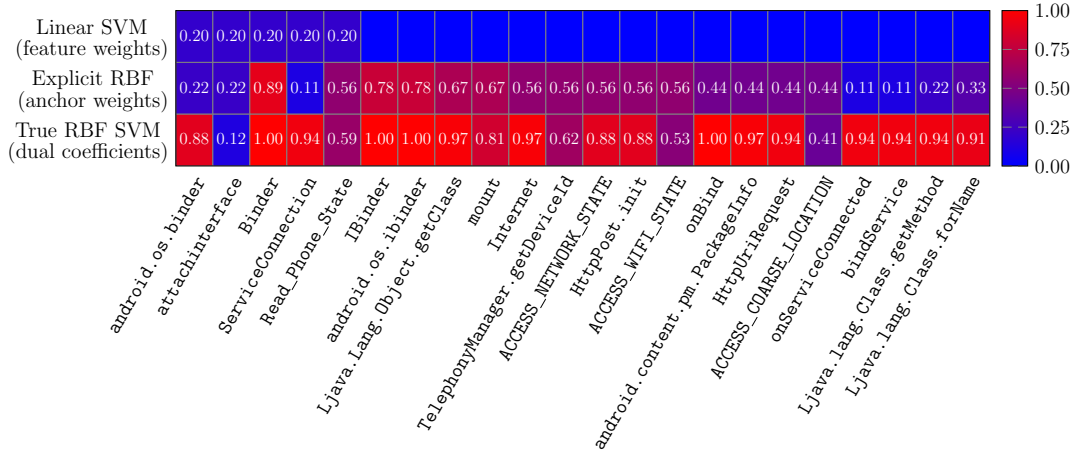

    \centering
    \input figures/hmap.tex
    \caption{Feature frequency heatmap for PBS across SVM formulations}
    \label{fig:frequency heatmap of PBS on SVM}
\end{figure}

The feature frequency heatmap in Figure~\ref{fig:frequency heatmap of PBS on SVM} 
shows how each SVM model reacts to bit-level changes. 
Note that the numbers in the heatmap are normalized frequencies---for each SVM variant, we run PBS 
for multiple iterations and at each step, selecting one feature's weight to flip. We count how many times 
each feature is selected and then normalize so that the most frequently selected feature has a value of~1.0, 
and everything else is scaled relative to that. For example, 0.5 means that feature was selected half 
as often as the most frequent one. Empty cells mean that feature was never selected by PBS for that 
model.

In the linear SVM, most of the perturbations 
are focused on a small number of features, particularly those related to inter-process communication 
and system-level interactions, such as \texttt{IBinder}, \texttt{Binder}, and \texttt{ServiceConnection}. 
These features appear repeatedly across iterations, indicating that the model relies heavily on a limited 
number of high-impact weights to define its decision boundary. As a result, modifying these features 
quickly leads to a sharp drop in accuracy. 

In the explicit RBF SVM, the perturbations are more distributed across features, including permissions 
such as \texttt{READ\un PHONE\un STATE} and network indicators like \texttt{INTERNET} 
and \texttt{ACCESS\un NETWORK\un STATE}. These features are commonly associated with 
malicious behavior, suggesting that the model captures a broader set of patterns compared to the 
linear SVM. However, certain features still appear more frequently than others, indicating that 
while PBS explores a larger set of parameters than in the linear SVM, some features remain 
disproportionately influential.

The true RBF SVM shows the most uniform distribution of feature influence. A wide range of features, 
including system calls, permissions, and API-related indicators such as \texttt{getClass}, \texttt{HttpPost}, 
and \texttt{ACCESS\un WIFI\un STATE} are involved in the perturbation process. This reflects the nature 
of the dual formulation, where the decision function is influenced by many support vectors rather than 
a small set of feature weights. As a result, no single feature dominates the model’s behavior, 
and perturbations are spread across a larger portion of the parameter space.

\subsection{MLP Results}

MLPs are neural models that are somewhat analogous to SVMs, and hence we train and test 
an MLP to serve as a comparison to our SVM results.
Note that the baseline accuracy for our MLP model is~0.9456.

\subsubsection{MLP Bit-Flip Procedure}

\begin{figure}[!htb]
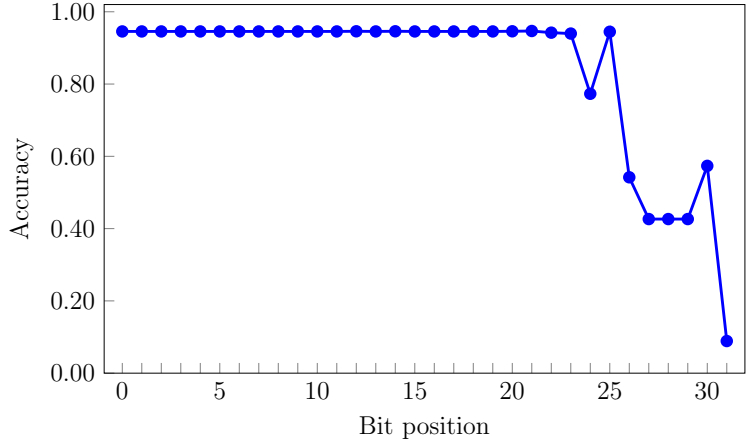

    \centering
    \input figures/lsb_mlp.tex
    \caption{Effect of LSB flips on MLP accuracy}
    \label{fig:LSB flips on MLP}
\end{figure}

For the MLP, the LSBs of the floating-point weights are flipped across all layers. 
For each experiment, the~$k \in \{1,2,\ldots,31\}$ LSBs (excluding sign bit) are flipped, the bit flip 
is applied across all weights in the network, and the model is re-evaluated on the 
validation set. This experiment produces the accuracy curve shown in Figure~\ref{fig:LSB flips on MLP}.

\subsubsection{Progressive Bit-Flip Search on Quantized MLP}

To evaluate parameter sensitivity in neural models, a PBS-inspired 
procedure is applied to our MLP model after quantizing its weights to an~8-bit representation. 
Following training in full precision, all weights across the network are scaled and converted 
to~8-bit integers. The objective of the attack is to identify and flip the bits that most significantly 
degrade model performance.

\begin{figure}[!htb]
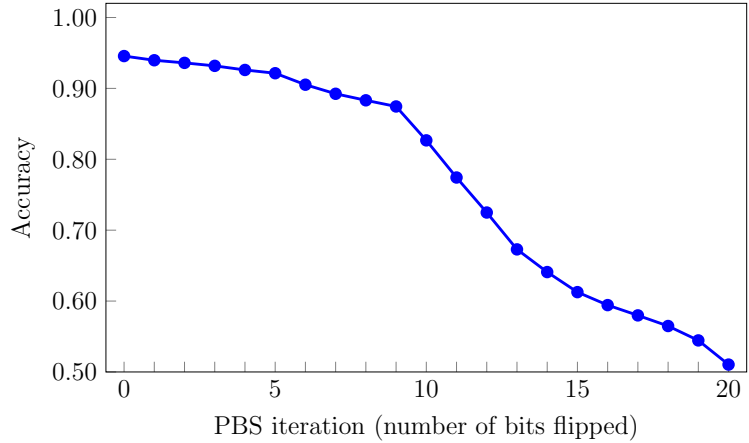

    \centering
    \input figures/pbs_mlp.tex
    \caption{Accuracy degradation of MLP under PBS across all layers}
    \label{fig:PBS on MLP}
\end{figure}

The results of the PBS attack on the MLP across all layers, as shown in Figure~\ref{fig:PBS on MLP}, 
indicate a gradual decrease in accuracy as the number of bit flips increases. The model maintains 
high accuracy during the initial iterations, with only a slight decline, and then degrades steadily 
as more bits are flipped. This behavior suggests that the MLP does not depend on a small set of 
critical parameters, but instead distributes importance across many weights.

\begin{figure}[!htb]
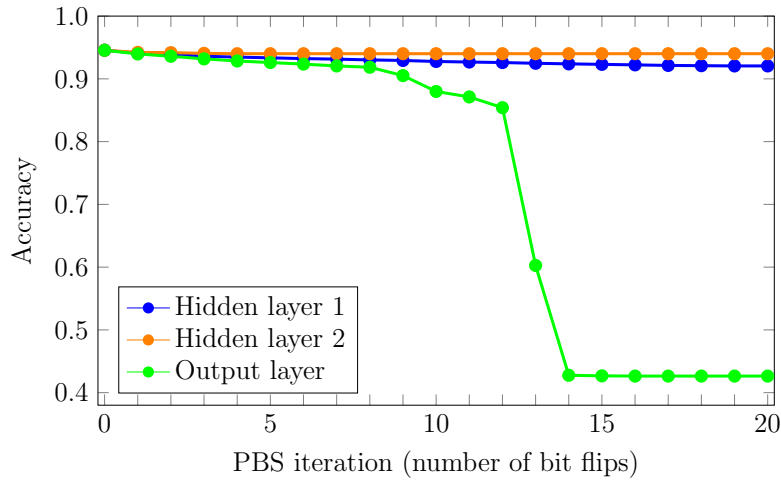

    \centering
    \input figures/pbs_mlp_layers.tex
    \caption{Accuracy degradation of MLP under PBS layer-wise}
    \label{fig:layerwise PBS MLP}
\end{figure}

In Figure~\ref{fig:layerwise PBS MLP}, we give the effect of PBS on individual layers,
where layer~1 is the first hidden layer, layer~2 is the second hidden layer, and layer~3
is the output layer.
We observe that the impact of perturbations varies significantly across the layers, 
with bit flips in the hidden layers resulting in only a minimal decrease in accuracy. 
In contrast, perturbations in the output layer cause rapid degradation, 
with accuracy dropping to near-random levels within the first~13 iterations.

These results indicate that parameter sensitivity in the MLP is highly unevenly distributed. 
The output layer is extremely fragile, while the hidden layers exhibits mild sensitivity. 
This suggests that critical decision-making parameters are concentrated in the output layer, 
while earlier layers contain more distributed and redundant representations.
This also suggests that an Extreme Learning Machine (ELM) model---where
hidden layer weights are assigned at random, and not adjusted during 
training---might be expected to perform well in this application~\cite{elm}.

\subsection{LSTM Results}

LSTMs are neural models that are somewhat analogous to HMMs, and hence we train and test 
an LSTM to serve as a comparison to our HMM results.
Note that the baseline accuracy for our LSTM model is~0.9480.

\subsubsection{LSTM Bit-Flip Procedure}

\begin{figure}[!htb]
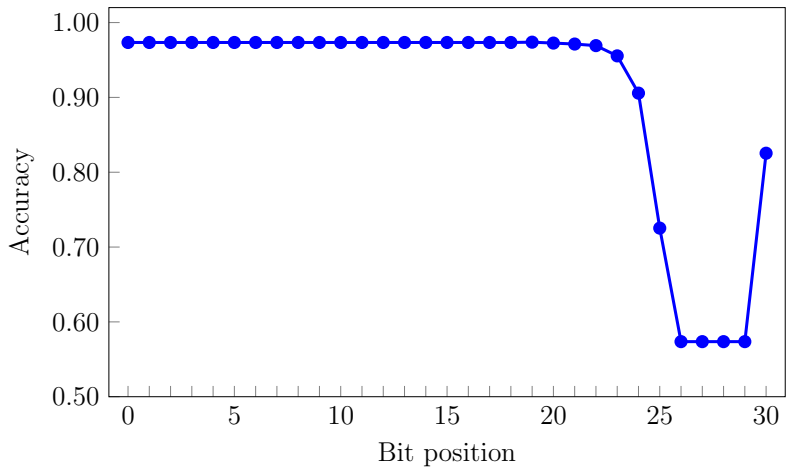

    \centering
    \input figures/lsb_lstm.tex
    \caption{Effect of LSB flips on LSTM accuracy}\label{fig:LSB on LSTM}
\end{figure}

For each experiment, $k \in \{1,2,\ldots,31\}$ LSBs (excluding sign bit) are flipped, the bit flip is applied 
across all weights in the network, and the LSTM model is re-evaluated on the validation set. 
This produces the accuracy curve shown in Figure~\ref{fig:LSB on LSTM}. The model exhibits
virtually no loss in accuracy until~22 bits are flipped, followed by a rapid decline to near-random 
accuracy levels. The slight increase in accuracy observed near~30 bits is likely due to the 
non-monotonic nature of LSB flipping, where flipping higher-order bits can partially cancel 
the effect of prior perturbations, resulting in a transient recovery before the model fully degrades.

\subsubsection{Progressive Bit-Flip Search on Quantized LSTM}

\begin{figure}[!htb]
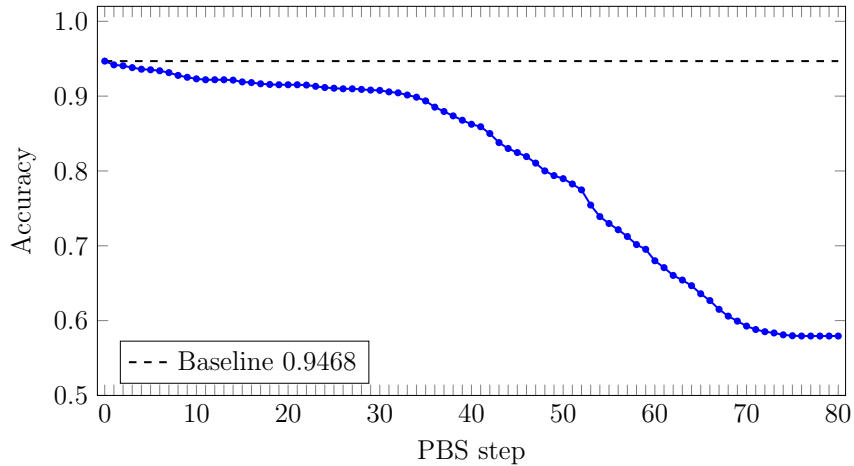

    \centering
    \input figures/pbs_lstm.tex
    \caption{Accuracy degradation of LSTM under PBS}
    \label{fig:Graph PBS on LSTM}
\end{figure}

To evaluate parameter sensitivity in sequential neural models, the PBS 
procedure is applied to the LSTM after quantizing its weights to an~8-bit representation. 
Following training in full precision, the model parameters are scaled and converted 
to~8-bit integers. The baseline validation accuracy of the LSTM is approximately~0.9500. As shown 
in Figure~\ref{fig:Graph PBS on LSTM}, the model maintains near-baseline performance 
during the initial iterations, with only minor decreases in accuracy. 

From Figure~\ref{fig:Graph PBS on LSTM}, we observe that
as more bits are flipped, 
the degradation becomes more pronounced, eventually stabilizing at around~0.58. 
This gradual decline indicates that the LSTM distributes importance across many 
parameters rather than relying on a small set of critical weights.

Overall, these results suggest that the LSTM is relatively robust to individual bit-level perturbations, 
but remains vulnerable under repeated targeted attacks. From a steganographic perspective, 
this distributed parameter structure allows modifications to be spread across a large number of parameters, 
enabling higher embedding capacity while maintaining acceptable performance.

\section{Conclusion and Future Work}\label{chap:conclusion}

In this chapter, we examined how classical machine learning models respond to controlled changes 
in their parameters. For HMM and SVM models, several types of perturbation were tested: 
direct bit flips on 32-bit floating point representations, 
8-bit quantization, targeted bit-flip search, and proportional probability adjustments. 
Analogous experiments on MLP and LSTM models were provided to give a broader comparison 
across model types.

These experiments highlighted that a limited set of parameters significantly influence model behavior. 
Classical models exhibited concentrated sensitivity, where a small number of parameters strongly 
affected performance, whereas neural models showed more distributed parameter importance. 
This difference directly impacts steganographic capacity, as classical models provide limited 
(but well-defined) regions for safe modification, while neural models allow changes to be spread 
across a larger number of parameters with less immediate impact on performance. From a practical standpoint, these findings suggest that deployed classical models may be more vulnerable to targeted parameter tampering than neural models, as a small number of bit flips can be sufficient to collapse their accuracy. Conversely, the larger and more distributed parameter spaces of neural models offer greater steganographic capacity, which has implications for both the security of deployed systems and the potential for covert communication through model parameters.

Future work can extend these results in several directions. Prior 
steganalysis work suggests that embedding data in neural networks often leaves detectable 
statistical bias in model parameters~\cite{JUST-2021-0197, Zhao2023CalibrationbasedSF}.
While our experiments show that meaningful parameter 
modifications are possible, it remains unclear whether such modifications are detectable
in the case of classical machine learning models.
Additionally, our analysis can be extended to other model families, such as 
Graph Neural Networks (GNN), where recent work has demonstrated vulnerability 
to bit-flip attacks and the need for specialized defense strategies~\cite{Kummer_2025}.
Finally, future work can explore training strategies or regularization techniques to improve 
robustness against parameter perturbations, as prior work has investigated methods for enhancing 
model stability under weight perturbations~\cite{tsai2021formalizinggeneralizationrobustnessneural}.

\bibliographystyle{plain}
\bibliography{references}

\end{document}

%% file: preamble.tex
\usepackage{amsmath,amsthm, amsfonts, amssymb, amsxtra, amsopn}
\usepackage{pgfplots}
\usepgfplotslibrary{colorbrewer}
\pgfplotsset{compat = 1.15, 
			 cycle list/Set1-3} 
\usetikzlibrary{pgfplots.statistics, pgfplots.colorbrewer} 
\usepackage{pgfplotstable}
\usepackage{graphicx,grffile}
\usepackage{multirow}
\usepackage{booktabs}
\usepackage{tcolorbox}
\usepackage{algorithm} 
\usepackage[noend]{algpseudocode} 
\usepackage{listings}
\usepackage{cmap}
\usepackage{colortbl}
\usepackage{adjustbox}
\usepackage{epsfig}

\usepackage[tableposition=top,font=small,skip=5pt,width=\textwidth]{caption}
\usepackage{subcaption}
\usepackage{makecell}

\usepackage[explicit]{titlesec}

\usetikzlibrary{patterns}

\def\bbb#1{{\color{blue}#1}}
\def\com#1{\bbb{\texttt{/\kern-1.5pt /} #1}}

\def\tau{\mathcal{T}}

\PassOptionsToPackage{hyphens}{url}
\PassOptionsToPackage{table}{xcolor}

\usepackage{hyperref}
\hypersetup{colorlinks=true,linkcolor=black,citecolor=black,urlcolor=blue,filecolor=black}
\hypersetup{pdfpagemode=UseNone,pdfstartview=}

\definecolor{darkgreen}{rgb}{0.125,0.5,0.169}

\usepackage[shortlabels]{enumitem}
\setlist[itemize]{noitemsep, topsep=0pt}

\advance\oddsidemargin by -0.5in
\advance\textwidth by 1.0in

\advance\topmargin by -0.5in
\advance\textheight by 1.0in

\long\def\symbolfootnotetext[#1]#2{\begingroup%
  \def\thefootnote{\fnsymbol{footnote}}\footnotetext[#1]{#2}\endgroup}

\newcommand\dunderline[3][-1pt]{{%
      \sbox0{#3}%
      \ooalign{\copy0\cr\rule[\dimexpr#1-#2\relax]{\wd0}{#2}}}}
\def\uuu{\kern-1pt\dunderline{0.75pt}{\phantom{M}}}

\DeclareMathOperator{\thth}{th}
\DeclareMathOperator{\mal}{mal}
\DeclareMathOperator{\ben}{ben}

\def\zz{\phantom{0}}

\def\un{\raisebox{1pt}{\underline{\phantom{M}}}}

%% file: figures/lsb_hmm_svm.tex
\begin{tikzpicture}[scale=0.90]
\begin{axis}[ 
		   width=11cm,
		   height=7.0cm,
		   scaled ticks=false,
	 	   x tick label style={scale=1.0,
   		 	/pgf/number format/.cd,
			/pgf/number format/1000 sep={},
   			fixed,
   			fixed zerofill,
    			precision=0,
			/tikz/.cd
		   },
		   x label style={scale=1.0},
	 	   y tick label style={scale=1.0,
        		 	/pgf/number format/.cd,
   			fixed,
   			fixed zerofill,
    			precision=1,
			/tikz/.cd
		    },
		   y label style={scale=1.0},
                    ymin=0.5, ymax=1.02,
                    symbolic x coords={1,2,3,4,5,6,7,8,9,10,11,12,13,14,15,16,17,18,19,20,21,22,23,24,25,26,27,28,29,30},
		    xtick=data,
                    xticklabels={1,,,,5,,,,,10,,,,,15,,,,,20,,,,,25,,,,,30},
                    ytick={0.5,0.6,0.7,0.8,0.9,1.0},
                    xlabel={Number of bits flipped},
                    ylabel={Accuracy},
                    enlarge x limits=0.01,
                    legend cell align=left,
                    legend pos=south west,
                    legend style={nodes={scale=1.0},
                         }
                    ] 
\addplot[forget plot,color=blue,very thick,mark=*,mark options={fill=blue, draw=blue}] coordinates { 
(1,0.9281)
(2,0.9281)
(3,0.9281)
(4,0.9281)
(5,0.9281)
(6,0.9281)
(7,0.9281)
(8,0.9281)
(9,0.9281)
(10,0.9281)
(11,0.9281)
(12,0.9281)
(13,0.9281)
(14,0.9281)
(15,0.9281)
(16,0.9281)
(17,0.9281)
(18,0.9281)
(19,0.9287)
(20,0.9287)
(21,0.9291)
(22,0.9291)
(23,0.9325)
(24,0.9376)
(25,0.6083)
(26,0.5619)
(27,0.5622)
(28,0.5534)
(29,0.5534)
(30,0.5534)
};
\addplot[forget plot,color=orange,very thick,mark=*,mark options={fill=orange, draw=orange}] coordinates { 
(1,0.989)
(2,0.989)
(3,0.989)
(4,0.989)
(5,0.989)
(6,0.989)
(7,0.989)
(8,0.989)
(9,0.989)
(10,0.989)
(11,0.989)
(12,0.989)
(13,0.989)
(14,0.989)
(15,0.989)
(16,0.989)
(17,0.989)
(18,0.989)
(19,0.9887)
(20,0.9877)
(21,0.988)
(22,0.9837)
(23,0.9638)
(24,0.9561)
(25,0.8371)
(26,0.9671)
(27,0.9791)
(28,0.9791)
(29,0.9791)
(30,0.5735)
};
\addlegendimage{mark=*, color=blue}
\addlegendentry{HMM}
\addlegendimage{mark=*, color=orange}
\addlegendentry{SVM}
\end{axis}
\end{tikzpicture}

%% file: figures/pbs_hmm.tex
\begin{tikzpicture}[scale=0.95, every node/.style={scale=0.90}]
\begin{axis}[
width=12cm,
height=7cm,
x tick label style={scale=0.9,
/pgf/number format/.cd,
/pgf/number format/1000 sep={},
fixed,
fixed zerofill,
precision=0,
/tikz/.cd
},
x label style={scale=1.0},
y tick label style={scale=1.0,
/pgf/number format/.cd,
fixed,
fixed zerofill,
precision=1,
/tikz/.cd
},
y label style={scale=1.0},
ymin=0.3, ymax=1.0,
symbolic x coords={A,B,C,D,E,F,G,H,I,J,K,L,M,N,O,P,Q,R,S,T,U,V,W,X,Y},
xtick=data,
xticklabels={0,1,2,3,4,5,6,7,8,9,10,11,12,13,14,15,16,17,18,19,20,21,22,23,24},
ytick={0.3,0.4,0.5,0.6,0.7,0.8,0.9,1.0},
xlabel={PBS iteration (number of bits flipped)},
ylabel={Validation accuracy},
enlarge x limits=0.01,
legend pos=north east,
legend style={nodes={scale=1.0},
}
]
\addplot[forget plot,color=black,very thick,no marks] coordinates {
(A,0.9137)
(B,0.8732)
(C,0.8452)
(D,0.8286)
(E,0.8128)
(F,0.7937)
(G,0.7784)
(H,0.7291)
(I,0.7146)
(J,0.7023)
(K,0.6908)
(L,0.6797)
(M,0.6699)
(N,0.6601)
(O,0.6533)
(P,0.6465)
(Q,0.6380)
(R,0.6316)
(S,0.6219)
(T,0.6125)
(U,0.5942)
(V,0.5861)
(W,0.5802)
(X,0.5734)
(Y,0.5691)
};
\addplot[forget plot,color=black,very thick,only marks,mark=*,mark options={fill=black, draw=black}] coordinates {
(A,0.9137)
};
\addplot[forget plot,color=green,very thick,only marks,mark=*,mark options={fill=green, draw=green}] coordinates {
(C,0.8452)
(G,0.7784)
(H,0.7291)
(J,0.7023)
(K,0.6908)
(P,0.6465)
(R,0.6316)
(S,0.6219)
(W,0.5802)
};
\addplot[forget plot,color=orange,very thick,only marks,mark=*,mark options={fill=orange, draw=orange}] coordinates {
(L,0.6797)
(M,0.6699)
(N,0.6601)
(O,0.6533)
(Q,0.6380)
(T,0.6125)
(U,0.5942)
(V,0.5861)
(X,0.5734)
};
\addplot[forget plot,color=red,very thick,only marks,mark=*,mark options={fill=red, draw=red}] coordinates {
(D,0.8286)
(F,0.7937)
(I,0.7146)
(Y,0.5691)
};
\addplot[forget plot,color=blue,very thick,only marks,mark=*,mark options={fill=blue, draw=blue}] coordinates {
(B,0.8732)
(E,0.8128)
};
\addlegendimage{only marks, mark=*, color=black}
\addlegendentry{initial}
\addlegendimage{only marks, mark=*, color=green}
\addlegendentry{$B_{\ben}$}
\addlegendimage{only marks, mark=*, color=orange}
\addlegendentry{$B_{\mal}$}
\addlegendimage{only marks, mark=*, color=red}
\addlegendentry{$A_{\mal}$}
\addlegendimage{only marks, mark=*, color=blue}
\addlegendentry{$A_{\ben}$}
\end{axis}
\end{tikzpicture}

%% file: figures/bar_all.tex
\begin{tikzpicture}[scale=0.9, every node/.style={scale=1.0}]
\pgfkeys{/pgf/number format/1000 sep={}}
\begin{axis}[
        width  = 13.25cm,
        height = 6.5cm,
        ymin=0.0,
        ymax=0.0655,
        ytick={0.0,0.01,0.02,0.03,0.04,0.05,0.06},
        scaled y ticks=false,
        major x tick style=transparent,
        ybar=5*\pgflinewidth,
        bar width=8pt,
        xlabel={PBS iteration},
        xlabel style={scale=1.0},
        ylabel={Accuracy drop},
        ylabel style={scale=1.0},
        symbolic x coords={
            1,2,3,4,5,6,7,8,9,10,11,12,
            13,14,15,16,17,18,19,20,21,22,23,24
        },
        xticklabels={
            1,2,3,4,5,6,7,8,9,10,11,12,
            13,14,15,16,17,18,19,20,21,22,23,24
        },
        xtick=data,
y tick label style={
    scale=0.9,
    /pgf/number format/fixed,
    /pgf/number format/fixed zerofill,
    /pgf/number format/precision=2
},
        x tick label style={
            scale=0.85,
            inner sep=0mm
        },
        nodes near coords,
every node near coord/.append style={
    rotate=90,
    scale=0.8,
    color=black,
    anchor=west,
    /pgf/number format/fixed,
    /pgf/number format/fixed zerofill,
    /pgf/number format/precision=4
},
        enlarge x limits=0.035,
        legend cell align=left,
        legend pos=north east,
        legend columns=1,
        legend style={nodes={scale=1.0}},
        every axis plot/.append style={
            ybar,
            bar width=8pt,
            bar shift=0pt,
            fill
        }
]

\addplot[forget plot, blue]
coordinates {(1,0.0370)};

\addplot[forget plot, green!70!black]
coordinates {(2,0.0281)};

\addplot[forget plot, red]
coordinates {(3,0.0166)};

\addplot[forget plot, blue]
coordinates {(4,0.0157)};

\addplot[forget plot, red]
coordinates {(5,0.0191)};

\addplot[forget plot, green!70!black]
coordinates {(6,0.0153)};

\addplot[forget plot, green!70!black]
coordinates {(7,0.0493)};

\addplot[forget plot, red]
coordinates {(8,0.0145)};

\addplot[forget plot, green!70!black]
coordinates {(9,0.0123)};

\addplot[forget plot, green!70!black]
coordinates {(10,0.0115)};

\addplot[forget plot, orange]
coordinates {(11,0.0111)};

\addplot[forget plot, orange]
coordinates {(12,0.0098)};

\addplot[forget plot, orange]
coordinates {(13,0.0098)};

\addplot[forget plot, orange]
coordinates {(14,0.0068)};

\addplot[forget plot, green!70!black]
coordinates {(15,0.0068)};

\addplot[forget plot, orange]
coordinates {(16,0.0085)};

\addplot[forget plot, green!70!black]
coordinates {(17,0.0064)};

\addplot[forget plot, green!70!black]
coordinates {(18,0.0098)};

\addplot[forget plot, orange]
coordinates {(19,0.0094)};

\addplot[forget plot, orange]
coordinates {(20,0.0183)};

\addplot[forget plot, orange]
coordinates {(21,0.0081)};

\addplot[forget plot, green!70!black]
coordinates {(22,0.0060)};

\addplot[forget plot, orange]
coordinates {(23,0.0068)};

\addplot[forget plot, red]
coordinates {(24,0.0043)};

\addlegendimage{area legend, fill=green!70!black, draw=green!70!black}
\addlegendentry{$B_{\ben}$}

\addlegendimage{area legend, fill=orange, draw=orange}
\addlegendentry{$B_{\mal}$}

\addlegendimage{area legend, fill=red, draw=red}
\addlegendentry{$A_{\mal}$}

\addlegendimage{area legend, fill=blue, draw=blue}
\addlegendentry{$A_{\ben}$}

\end{axis}
\end{tikzpicture}

%% file: figures/sense_ben_hmm.tex
\begin{tikzpicture}[scale=0.9]
\begin{axis}[ 
		   width=14cm,
		   height=7cm,
		   scaled ticks=false,
	 	   x tick label style={scale=0.9,
   		 	/pgf/number format/.cd,
			/pgf/number format/1000 sep={},
   			fixed,
   			fixed zerofill,
    			precision=0,
			/tikz/.cd
		   },
		   x label style={scale=1.0},
	 	   y tick label style={scale=0.9,
        		 	/pgf/number format/.cd,
   			fixed,
   			fixed zerofill,
    			precision=3,
			/tikz/.cd
		    },
		   y label style={scale=1.0},
                    ymin=0.0, ymax=0.016,
                    symbolic x coords={A,B,C,D,E},
		    xtick=data,
                    xticklabels={5\%,10\%,20\%,50\%,75\%},
                    ytick={0.000,0.002,0.004,0.006,0.008,0.010,0.012,0.014,0.016},
                    xlabel={Perturbation level},
                    ylabel={$\Delta$ score},
                    enlarge x limits=0.01,
                    legend cell align=left,
                    legend pos=north west,
                    legend style={nodes={scale=0.8},
                         }
                    ] 
\addplot[forget plot,color=blue,very thick,mark=*,mark options={fill=blue, draw=blue}] coordinates { 
(A,0.000362)
(B,0.000771)
(C,0.001751)
(D,0.006657)
(E,0.015823)
};
\addplot[forget plot,color=orange,very thick,mark=*,mark options={fill=orange, draw=orange}] coordinates { 
(A,0.000333)
(B,0.000704)
(C,0.001581)
(D,0.005842)
(E,0.01365)
};
\addplot[forget plot,color=green,very thick,mark=*,mark options={fill=green, draw=green}] coordinates { 
(A,0.000374)
(B,0.000778)
(C,0.001697)
(D,0.005773)
(E,0.012577)
};
\addplot[forget plot,color=red,very thick,mark=*,mark options={fill=red, draw=red}] coordinates { 
(A,0.000348)
(B,0.000727)
(C,0.001597)
(D,0.005555)
(E,0.012354)
};
\addplot[forget plot,color=purple,very thick,mark=*,mark options={fill=purple, draw=purple}] coordinates { 
(A,0.000349)
(B,0.000729)
(C,0.001599)
(D,0.005527)
(E,0.012162)
};
\addplot[forget plot,color=brown,very thick,mark=*,mark options={fill=brown, draw=brown}] coordinates { 
(A,0.000352)
(B,0.000734)
(C,0.001604)
(D,0.005483)
(E,0.011924)
};
\addlegendimage{mark=*, color=blue}
\addlegendentry{\texttt{Landroid.content.Context.registerReceiver}}
\addlegendimage{mark=*, color=orange}
\addlegendentry{\texttt{Landroid.content.Context.unregisterReceiver}}
\addlegendimage{mark=*, color=green}
\addlegendentry{\texttt{ServiceConnection}}
\addlegendimage{mark=*, color=red}
\addlegendentry{\texttt{onServiceConnected}}
\addlegendimage{mark=*, color=purple}
\addlegendentry{\texttt{android.os.Binder}}
\addlegendimage{mark=*, color=brown}
\addlegendentry{\texttt{bindService}}
\end{axis}
\end{tikzpicture}

%% file: figures/sense_mal_hmm.tex
\begin{tikzpicture}[scale=0.9]
\begin{axis}[ 
		   width=14cm,
		   height=7cm,
		   scaled ticks=false,
	 	   x tick label style={scale=0.9,
   		 	/pgf/number format/.cd,
			/pgf/number format/1000 sep={},
   			fixed,
   			fixed zerofill,
    			precision=0,
			/tikz/.cd
		   },
		   x label style={scale=1.0},
	 	   y tick label style={scale=0.9,
        		 	/pgf/number format/.cd,
   			fixed,
   			fixed zerofill,
    			precision=3,
			/tikz/.cd
		    },
		   y label style={scale=1.0},
                    ymin=0.0, ymax=0.032,
                    symbolic x coords={A,B,C,D,E},
		    xtick=data,
                    xticklabels={5\%,10\%,20\%,50\%,75\%},
                    ytick={0.000,0.005,0.010,0.015,0.020,0.025,0.030},
                    xlabel={Perturbation level},
                    ylabel={$\Delta$ score},
                    enlarge x limits=0.01,
                    legend cell align=left,
                    legend pos=north west,
                    legend style={nodes={scale=0.8},
                         }
                    ] 
\addplot[forget plot,color=blue,very thick,mark=*,mark options={fill=blue, draw=blue}] coordinates { 
(A,0.00067)
(B,0.001431)
(C,0.003268)
(D,0.012697)
(E,0.030931)
};
\addplot[forget plot,color=orange,very thick,mark=*,mark options={fill=orange, draw=orange}] coordinates { 
(A,0.000668)
(B,0.001398)
(C,0.003069)
(D,0.010638)
(E,0.02342)
};
\addplot[forget plot,color=green,very thick,mark=*,mark options={fill=green, draw=green}] coordinates { 
(A,0.000396)
(B,0.00086)
(C,0.002022)
(D,0.008402)
(E,0.02135)
};
\addplot[forget plot,color=red,very thick,mark=*,mark options={fill=red, draw=red}] coordinates { 
(A,0.000328)
(B,0.000715)
(C,0.001698)
(D,0.00719)
(E,0.018412)
};
\addplot[forget plot,color=purple,very thick,mark=*,mark options={fill=purple, draw=purple}] coordinates { 
(A,0.000372)
(B,0.000782)
(C,0.00174)
(D,0.006264)
(E,0.014304)
};
\addplot[forget plot,color=brown,very thick,mark=*,mark options={fill=brown, draw=brown}] coordinates { 
(A,0.000556)
(B,0.001129)
(C,0.002315)
(D,0.006112)
(E,0.009445)
};
\addlegendimage{mark=*, color=blue}
\addlegendentry{\texttt{INTERNET}}
\addlegendimage{mark=*, color=orange}
\addlegendentry{\texttt{READ\un PHONE\un STATE}}
\addlegendimage{mark=*, color=green}
\addlegendentry{\texttt{WRITE\un EXTERNAL\un STORAGE}}
\addlegendimage{mark=*, color=red}
\addlegendentry{\texttt{HttpUriRequest}}
\addlegendimage{mark=*, color=purple}
\addlegendentry{\texttt{android.intent.action.BOOT\un COMPLETED}}
\addlegendimage{mark=*, color=brown}
\addlegendentry{\texttt{attachInterface}}
\end{axis}
\end{tikzpicture}

%% file: figures/pbs_linear_svm.tex
\begin{tikzpicture}[scale=0.95, every node/.style={scale=0.90}]
\begin{axis}[ 
		   width=12cm,
		   height=7cm,
		   scaled ticks=false,
	 	   x tick label style={scale=1.0,
   		 	/pgf/number format/.cd,
			/pgf/number format/1000 sep={},
   			fixed,
   			fixed zerofill,
    			precision=0,
			/tikz/.cd
		   },
		   x label style={scale=1.0},
	 	   y tick label style={scale=1.0,
        		 	/pgf/number format/.cd,
   			fixed,
   			fixed zerofill,
    			precision=1,
			/tikz/.cd
		    },
		   y label style={scale=1.0},
                    ymin=0.5, ymax=1.02,
                    symbolic x coords={0,1,2,3,4,5},
		    xtick=data,
                    xticklabels={0,1,2,3,4,5},
                    ytick={0.5,0.6,0.7,0.8,0.9,1.0},
                    xlabel={PBS step},
                    ylabel={Accuracy},
                    enlarge x limits=0.01,
                    legend cell align=left,
                    legend pos=south west,
                    legend style={nodes={scale=1.0},
                         }
                    ] 
\addplot[forget plot,color=black,thick,dashed, no marks] coordinates { 
(0, 0.9871)
(1, 0.9871)
(2, 0.9871)
(3, 0.9871)
(4, 0.9871)
(5, 0.9871)
};
\addplot[forget plot,color=blue,very thick,mark=*,mark options={fill=blue, draw=blue}] coordinates { 
(0, 0.9871)
(1, 0.9418)
(2, 0.7947)
(3, 0.618)
(4, 0.5761)
(5, 0.5736)
};
\addlegendimage{thick, dashed, color=black}
\addlegendentry{Baseline 0.9871}
\end{axis}
\end{tikzpicture}

%% file: figures/pbs_svm_all.tex
\begin{tikzpicture}[scale=0.95, every node/.style={scale=0.90}]
\begin{axis}[ 
		   width=12cm,
		   height=7cm,
		   scaled ticks=false,
	 	   x tick label style={scale=1.0,
   		 	/pgf/number format/.cd,
			/pgf/number format/1000 sep={},
   			fixed,
   			fixed zerofill,
    			precision=0,
			/tikz/.cd
		   },
		   x label style={scale=1.0},
	 	   y tick label style={scale=1.0,
        		 	/pgf/number format/.cd,
   			fixed,
   			fixed zerofill,
    			precision=1,
			/tikz/.cd
		    },
		   y label style={scale=1.0},
                    ymin=0.4, ymax=1.02,
                    symbolic x coords={0,1,2,3,4,5,6,7,8,9,10,11,12,13,14,15,16,17,18,19,20,21,22,23,24,25,26,27,28,29,30,31,32},
		    xtick=data,
                    xticklabels={0,,,,,5,,,,,10,,,,,15,,,,,20,,,,,25,,,,,30},
                    ytick={0.4,0.5,0.6,0.7,0.8,0.9,1.0},
                    xlabel={PBS iteration (number of bit flips)},
                    ylabel={Accuracy},
                    enlarge x limits=0.01,
                    legend cell align=left,
                    legend pos=north east,
                    legend style={nodes={scale=1.0},
                         }
                    ] 
\addplot[forget plot,color=red,very thick,mark=*,mark options={fill=red, draw=red}] coordinates { 
(0, 0.9871)
(1, 0.9418)
(2, 0.7947)
(3, 0.618)
(4, 0.5761)
(5, 0.5736)
(6, 0.5736)
(7, 0.5736)
(8, 0.5736)
(9, 0.5736)
(10, 0.5736)
(11, 0.5736)
(12, 0.5736)
(13, 0.5736)
(14, 0.5736)
(15, 0.5736)
(16, 0.5736)
(17, 0.5736)
(18, 0.5736)
(19, 0.5736)
(20, 0.5736)
};
\addplot[forget plot,color=orange,very thick,mark=*,mark options={fill=orange, draw=orange}] coordinates { 
(0, 0.9867)
(1, 0.9717)
(2, 0.9057)
(3, 0.7174)
(4, 0.6097)
(5, 0.5237)
(6, 0.4593)
(7, 0.4451)
(8, 0.4397)
(9, 0.4293)
(10, 0.4293)
(11, 0.4293)
(12, 0.4293)
(13, 0.4293)
(14, 0.4293)
(15, 0.4293)
(16, 0.4293)
(17, 0.4293)
(18, 0.4293)
(19, 0.4293)
(20, 0.4293)
};
\addplot[forget plot,color=blue,very thick,mark=*,mark options={fill=blue, draw=blue}] coordinates { 
(0, 0.9921)
(1, 0.9909)
(2, 0.9904)
(3, 0.9892)
(4, 0.9875)
(5, 0.9863)
(6, 0.9855)
(7, 0.9821)
(8, 0.9626)
(9, 0.9431)
(10, 0.9202)
(11, 0.8869)
(12, 0.8317)
(13, 0.786)
(14, 0.7319)
(15, 0.6825)
(16, 0.6505)
(17, 0.6351)
(18, 0.623)
(19, 0.6122)
(20, 0.606)
(21, 0.6002)
(22, 0.5948)
(23, 0.5914)
(24, 0.5898)
(25, 0.5885)
(26, 0.5877)
(27, 0.586)
(28, 0.5852)
(29, 0.5827)
(30, 0.5819)
(31, 0.581)
(32, 0.5806)
};
\addlegendimage{mark=*, color=red}
\addlegendentry{Linear SVM}
\addlegendimage{mark=*, color=orange}
\addlegendentry{Explicit RBF SVM}
\addlegendimage{mark=*, color=blue}
\addlegendentry{True RBF SVM}
\end{axis}
\end{tikzpicture}

%% file: figures/hmap.tex
\begin{tikzpicture}[scale=0.85]
    \begin{axis}[
        width=14.25cm,
        height=4.0cm,
	colormap={redblue}{color=(blue) color=(red)},
        xticklabels={
\texttt{android.os.binder},
\texttt{attachinterface},
\texttt{Binder},
\texttt{ServiceConnection},
\texttt{Read\un Phone\un State},
\texttt{IBinder},
\texttt{android.os.ibinder},
\texttt{Ljava.Lang.Object.getClass},
\texttt{mount},
\texttt{Internet},
\texttt{TelephonyManager.getDeviceId},
\texttt{ACCESS\un NETWORK\un STATE},
\texttt{HttpPost.init},
\texttt{ACCESS\un WIFI\un STATE},
\texttt{onBind},
\texttt{android.content.pm.PackageInfo},
\texttt{HttpUriRequest},
\texttt{ACCESS\un COARSE\un LOCATION},
\texttt{onServiceConnected},
\texttt{bindService},
\texttt{Ljava.lang.Class.getMethod},
\texttt{Ljava.lang.Class.forName},
        },
        xtick={0,...,21},
        xtick style={draw=none},
	xticklabel style={anchor=east,rotate=60,yshift=-5pt,scale=0.75},
        yticklabel style={scale=0.785, align=center},
        yticklabels={
\raisebox{-2.25ex}{Linear SVM}\\ \raisebox{-1.75ex}{(feature weights)},
\raisebox{-2.25ex}{Explicit RBF}\\ \raisebox{-1.75ex}{(anchor weights)},
\raisebox{-2.25ex}{True RBF SVM}\\ \raisebox{-1.75ex}{(dual coefficients)},
        },
        ytick={0,...,2},
        ytick style={draw=none},
        enlargelimits=false,
        colorbar,
        colorbar style={xshift=-5pt, width=4mm,
            ytick={0.00,0.25,0.50,0.75,1.00},
            yticklabels={0.00,0.25,0.50,0.75,1.00},
            yticklabel={\pgfmathprintnumber\tick},
            yticklabel style={
            		scale=0.75,
            		/pgf/number format/fixed,
            		/pgf/number format/fixed zerofill,
			/pgf/number format/precision=2}
        },
        point meta min=0.0,
        point meta max=1.0,
        nodes near coords={\pgfmathprintnumber\pgfplotspointmeta},
        nodes near coords black white/.style={
            small value/.style={
                yshift=-4.5pt,
                text=black,
                /pgf/number format/fixed,
                /pgf/number format/precision=2,
                /pgf/number format/zerofill=true,
                scale=0.625,
            },
            large value/.style={
                yshift=-4.5pt,
                text=white,
                /pgf/number format/fixed,
                /pgf/number format/precision=2,
                /pgf/number format/zerofill=true,
                scale=0.625,
            },
            every node near coord/.style={
                check for zero/.code={
                    \pgfmathfloatifflags{\pgfplotspointmeta}{0}{
                        \pgfkeys{/tikz/coordinate}
                    }{
                        \begingroup
                        \pgfkeys{/pgf/fpu}
                        \pgfmathparse{\pgfplotspointmeta<#1}
                        \global\let\result=\pgfmathresult
                        \endgroup
                        %
                        %
                        \pgfmathfloatcreate{1}{1.0}{0}
                        \let\ONE=\pgfmathresult
                        \ifx\result\ONE
                            \pgfkeysalso{/pgfplots/small value}
                        \else
                            \pgfkeysalso{/pgfplots/large value}
                        \fi
                    }
                },
                check for zero,
            },
        },
        nodes near coords black white=0.1,
    ]
        \addplot[
            matrix plot,
            mesh/cols=22,
            point meta=explicit,draw=gray
        ] table [meta=C] {
            x y C
0 0 0.20
1 0 0.20
2 0 0.20
3 0 0.20
4 0 0.20
5 0 0.0
6 0 0.0
7 0 0.0
8 0 0.0
9 0 0.0
10 0 0.0
11 0 0.0
12 0 0.0
13 0 0.0
14 0 0.0
15 0 0.0
16 0 0.0
17 0 0.0
18 0 0.0
19 0 0.0
20 0 0.0
21 0 0.0
0 1 0.22
1 1 0.22
2 1 0.89
3 1 0.11
4 1 0.56
5 1 0.78
6 1 0.78
7 1 0.67
8 1 0.67
9 1 0.56
10 1 0.56
11 1 0.56
12 1 0.56
13 1 0.56
14 1 0.44
15 1 0.44
16 1 0.44
17 1 0.44
18 1 0.11
19 1 0.11
20 1 0.22
21 1 0.33
0 2 0.88
1 2 0.12
2 2 1.00
3 2 0.94
4 2 0.59
5 2 1.00
6 2 1.00
7 2 0.97
8 2 0.81
9 2 0.97
10 2 0.62
11 2 0.88
12 2 0.88
13 2 0.53
14 2 1.00
15 2 0.97
16 2 0.94
17 2 0.41
18 2 0.94
19 2 0.94
20 2 0.94
21 2 0.91
         };
    \end{axis}
\end{tikzpicture}

%% file: figures/lsb_mlp.tex
\begin{tikzpicture}[scale=0.9, every node/.style={scale=0.9}]
\begin{axis}[
		   width=11cm,
		   height=7cm,
                    symbolic x coords={0,1,2,3,4,5,6,7,8,9,10,11,12,13,14,15,16,17,18,19,20,21,22,23,24,25,26,27,28,29,30,31,32},
                    xtick={0,1,2,3,4,5,6,7,8,9,10,11,12,13,14,15,16,17,18,19,20,21,22,23,24,25,26,27,28,29,30,31,32},
                    xticklabels={0,,,,,5,,,,,10,,,,,15,,,,,20,,,,,25,,,,,30},
	 	   x tick label style={
		   },
		   xtick pos=bottom,
		   ytick pos=left,
	 	   y tick label style={
    		 	/pgf/number format/.cd,
   			fixed,
   			fixed zerofill,
    			precision=2,
			/tikz/.cd},
		   ytick={0.0,0.2,0.4,0.6,0.8,1.0},
  		   scaled y ticks=false,
		   enlarge x limits=0.03,
                    ymin=0.0,
                    ymax=1.02,
                    xlabel={Bit position},
                    ylabel={Accuracy}] 
\addplot[color=blue,very thick,mark=*] coordinates {
(0, 0.9456)
(1, 0.9456)
(2, 0.9456)
(3, 0.9456)
(4, 0.9456)
(5, 0.9456)
(6, 0.9456)
(7, 0.9456)
(8, 0.9456)
(9, 0.9456)
(10, 0.9456)
(11, 0.9456)
(12, 0.946)
(13, 0.9456)
(14, 0.946)
(15, 0.9456)
(16, 0.9456)
(17, 0.9456)
(18, 0.9456)
(19, 0.9456)
(20, 0.946)
(21, 0.9468)
(22, 0.9422)
(23, 0.9397)
(24, 0.7731)
(25, 0.9447)
(26, 0.542)
(27, 0.4264)
(28, 0.4264)
(29, 0.4264)
(30, 0.5736)
(31, 0.0889)
};
\end{axis}
\end{tikzpicture}

%% file: figures/pbs_mlp.tex
\begin{tikzpicture}[scale=0.9, every node/.style={scale=0.9}]
\begin{axis}[
		   width=11cm,
		   height=7cm,
                    symbolic x coords={0,1,2,3,4,5,6,7,8,9,10,11,12,13,14,15,16,17,18,19,20},
                    xtick={0,1,2,3,4,5,6,7,8,9,10,11,12,13,14,15,16,17,18,19,20},
                    xticklabels={0,,,,,5,,,,,10,,,,,15,,,,,20},
	 	   x tick label style={
		   },
		   xtick pos=bottom,
		   ytick pos=left,
	 	   y tick label style={
    		 	/pgf/number format/.cd,
   			fixed,
   			fixed zerofill,
    			precision=2,
			/tikz/.cd},
		   ytick={0.5,0.6,0.7,0.8,0.9,1.0},
  		   scaled y ticks=false,
		   enlarge x limits=0.03,
                    ymin=0.5,
                    ymax=1.02,
                    xlabel={PBS iteration (number of bits flipped)},
                    ylabel={Accuracy}] 
\addplot[color=blue,very thick,mark=*] coordinates {
(0, 0.9456)
(1, 0.9397)
(2, 0.936)
(3, 0.9318)
(4, 0.926)
(5, 0.9214)
(6, 0.9052)
(7, 0.8924)
(8, 0.8832)
(9, 0.8745)
(10, 0.8267)
(11, 0.7743)
(12, 0.7249)
(13, 0.6729)
(14, 0.6409)
(15, 0.6126)
(16, 0.5943)
(17, 0.5798)
(18, 0.5648)
(19, 0.5445)
(20, 0.5104)
};
\end{axis}
\end{tikzpicture}

%% file: figures/pbs_mlp_layers.tex
\begin{tikzpicture}[scale=0.95, every node/.style={scale=0.90}]
\begin{axis}[ 
		   width=11cm,
		   height=7cm,
		   scaled ticks=false,
	 	   x tick label style={scale=1.0,
   		 	/pgf/number format/.cd,
			/pgf/number format/1000 sep={},
   			fixed,
   			fixed zerofill,
    			precision=0,
			/tikz/.cd
		   },
		   x label style={scale=1.0},
	 	   y tick label style={scale=1.0,
        		 	/pgf/number format/.cd,
   			fixed,
   			fixed zerofill,
    			precision=1,
			/tikz/.cd
		    },
		   y label style={scale=1.0},
                    ymin=0.38, ymax=1.0,
                    symbolic x coords={0,1,2,3,4,5,6,7,8,9,10,11,12,13,14,15,16,17,18,19,20},
		    xtick=data,
                    xticklabels={0,,,,,5,,,,,10,,,,,15,,,,,20},
                    ytick={0.4,0.5,0.6,0.7,0.8,0.9,1.0},
                    xlabel={PBS iteration (number of bit flips)},
                    ylabel={Accuracy},
                    enlarge x limits=0.01,
                    legend cell align=left,
                    legend pos=south west,
                    legend style={nodes={scale=1.0},
                         }
                    ] 
\addplot[forget plot,color=blue,very thick,mark=*,mark options={fill=blue, draw=blue}] coordinates { 
(0, 0.9456)
(1, 0.941)
(2, 0.9389)
(3, 0.936)
(4, 0.9347)
(5, 0.9335)
(6, 0.9323)
(7, 0.9314)
(8, 0.9302)
(9, 0.9293)
(10, 0.9277)
(11, 0.9268)
(12, 0.926)
(13, 0.9248)
(14, 0.9239)
(15, 0.9231)
(16, 0.9223)
(17, 0.9214)
(18, 0.921)
(19, 0.9206)
(20, 0.9206)
};
\addplot[forget plot,color=orange,very thick,mark=*,mark options={fill=orange, draw=orange}] coordinates { 
(0, 0.9456)
(1, 0.9422)
(2, 0.9418)
(3, 0.9406)
(4, 0.9401)
(5, 0.9401)
(6, 0.9401)
(7, 0.9401)
(8, 0.9401)
(9, 0.9401)
(10, 0.9401)
(11, 0.9401)
(12, 0.9401)
(13, 0.9401)
(14, 0.9401)
(15, 0.9401)
(16, 0.9401)
(17, 0.9401)
(18, 0.9401)
(19, 0.9401)
(20, 0.9401)
};
\addplot[forget plot,color=green,very thick,mark=*,mark options={fill=green, draw=green}] coordinates { 
(0, 0.9456)
(1, 0.9397)
(2, 0.936)
(3, 0.9318)
(4, 0.9285)
(5, 0.926)
(6, 0.9235)
(7, 0.9206)
(8, 0.9185)
(9, 0.9052)
(10, 0.8799)
(11, 0.8712)
(12, 0.8541)
(13, 0.6027)
(14, 0.4277)
(15, 0.4268)
(16, 0.4264)
(17, 0.4264)
(18, 0.4264)
(19, 0.4264)
(20, 0.4264)
};
\addlegendimage{mark=*, color=blue}
\addlegendentry{Hidden layer~1}
\addlegendimage{mark=*, color=orange}
\addlegendentry{Hidden layer~2}
\addlegendimage{mark=*, color=green}
\addlegendentry{Output layer}
\end{axis}
\end{tikzpicture}

%% file: figures/lsb_lstm.tex
\begin{tikzpicture}[scale=0.95, every node/.style={scale=0.9}]
\begin{axis}[
		   width=11cm,
		   height=7cm,
                    symbolic x coords={0,1,2,3,4,5,6,7,8,9,10,11,12,13,14,15,16,17,18,19,20,21,22,23,24,25,26,27,28,29,30},
                    xtick={0,1,2,3,4,5,6,7,8,9,10,11,12,13,14,15,16,17,18,19,20,21,22,23,24,25,26,27,28,29,30},
                    xticklabels={0,,,,,5,,,,,10,,,,,15,,,,,20,,,,,25,,,,,30},
	 	   x tick label style={
		   },
		   xtick pos=bottom,
		   ytick pos=left,
	 	   y tick label style={
    		 	/pgf/number format/.cd,
   			fixed,
   			fixed zerofill,
    			precision=2,
			/tikz/.cd},
		   ytick={0.5,0.6,0.7,0.8,0.9,1.0},
  		   scaled y ticks=false,
		   enlarge x limits=0.03,
                    ymin=0.5,
                    ymax=1.02,
                    xlabel={Bit position},
                    ylabel={Accuracy}] 
\addplot[color=blue,very thick,mark=*] coordinates {
(0, 0.9734)
(1, 0.9734)
(2, 0.9734)
(3, 0.9734)
(4, 0.9734)
(5, 0.9734)
(6, 0.9734)
(7, 0.9734)
(8, 0.9734)
(9, 0.9734)
(10, 0.9734)
(11, 0.9734)
(12, 0.9734)
(13, 0.9734)
(14, 0.9734)
(15, 0.9734)
(16, 0.9734)
(17, 0.9734)
(18, 0.9734)
(19, 0.9738)
(20, 0.9726)
(21, 0.9713)
(22, 0.9692)
(23, 0.9555)
(24, 0.9057)
(25, 0.7253)
(26, 0.5736)
(27, 0.5736)
(28, 0.5736)
(29, 0.5736)
(30, 0.8254)
};
\end{axis}
\end{tikzpicture}

%% file: figures/pbs_lstm.tex
\begin{tikzpicture}[scale=0.95, every node/.style={scale=0.90}]
\begin{axis}[ 
		   width=12cm,
		   height=7.0cm,
		   scaled ticks=false,
	 	   x tick label style={scale=1.0,
   		 	/pgf/number format/.cd,
			/pgf/number format/1000 sep={},
   			fixed,
   			fixed zerofill,
    			precision=0,
			/tikz/.cd
		   },
		   x label style={scale=1.0},
	 	   y tick label style={scale=1.0,
        		 	/pgf/number format/.cd,
   			fixed,
   			fixed zerofill,
    			precision=1,
			/tikz/.cd
		    },
		   y label style={scale=1.0},
                    ymin=0.5, ymax=1.02,
		   xtick=data,
                    xticklabels={0,10,20,30,40,50,60,70,80},
                    ytick={0.5,0.6,0.7,0.8,0.9,1.0},
                    xlabel={PBS step},
                    ylabel={Accuracy},
                    enlarge x limits=0.01,
                    legend cell align=left,
                    legend pos=south west,
                    legend style={nodes={scale=1.0},
                         }
                    ] 
\addplot[forget plot,color=black,thick,dashed, no marks] coordinates { 
(0, 0.9468)
(10, 0.9468)
(20, 0.9468)
(30, 0.9468)
(40, 0.9468)
(50, 0.9468)
(60, 0.9468)
(70, 0.9468)
(80, 0.9468)
};
\addplot[forget plot,color=blue,thick,mark=*,mark size=1.0pt, mark options={fill=blue, draw=blue}] coordinates { 
(0, 0.9468)
(1, 0.9418)
(2, 0.9406)
(3, 0.9381)
(4, 0.936)
(5, 0.9352)
(6, 0.9339)
(7, 0.9314)
(8, 0.9277)
(9, 0.9252)
(10, 0.9231)
(11, 0.9219)
(12, 0.9219)
(13, 0.9219)
(14, 0.9214)
(15, 0.919)
(16, 0.9181)
(17, 0.9165)
(18, 0.9156)
(19, 0.9152)
(20, 0.9152)
(21, 0.9152)
(22, 0.9148)
(23, 0.9131)
(24, 0.9115)
(25, 0.9106)
(26, 0.9098)
(27, 0.9098)
(28, 0.909)
(29, 0.9081)
(30, 0.9077)
(31, 0.9057)
(32, 0.9044)
(33, 0.9015)
(34, 0.8986)
(35, 0.8936)
(36, 0.8853)
(37, 0.8795)
(38, 0.8736)
(39, 0.8678)
(40, 0.8624)
(41, 0.8591)
(42, 0.85)
(43, 0.8379)
(44, 0.83)
(45, 0.8246)
(46, 0.8192)
(47, 0.8105)
(48, 0.8001)
(49, 0.7938)
(50, 0.7897)
(51, 0.7826)
(52, 0.7747)
(53, 0.7544)
(54, 0.739)
(55, 0.7298)
(56, 0.7215)
(57, 0.7124)
(58, 0.7016)
(59, 0.6953)
(60, 0.68)
(61, 0.6708)
(62, 0.6604)
(63, 0.6542)
(64, 0.6467)
(65, 0.6359)
(66, 0.6268)
(67, 0.6151)
(68, 0.606)
(69, 0.5993)
(70, 0.5927)
(71, 0.5881)
(72, 0.5852)
(73, 0.5835)
(74, 0.581)
(75, 0.5798)
(76, 0.5794)
(77, 0.5794)
(78, 0.5794)
(79, 0.5794)
(80, 0.5794)
};
\addlegendimage{thick, dashed, color=black}
\addlegendentry{Baseline 0.9468}
\end{axis}
\end{tikzpicture}